\documentclass[10pt,twocolumn,letterpaper]{article}
\usepackage{cvpr}              
\usepackage[accsupp]{axessibility}
\usepackage[T1]{fontenc}

\usepackage{amsmath,amssymb,amsfonts}
\usepackage{setspace}
\usepackage{array}
\usepackage{algorithmic}
\usepackage{algorithm}
\usepackage{graphicx}
\usepackage{textcomp}
\usepackage{booktabs}
\usepackage{ragged2e}
\usepackage{diagbox}
\usepackage{multirow}
\usepackage{makecell}
\usepackage[table]{xcolor}
\newcommand{\best}[1]{\cellcolor{gray!20}\textbf{#1}}
\usepackage{colortbl}
\usepackage{enumitem}
\usepackage{dsfont}
\usepackage[T1]{fontenc}
\usepackage{inconsolata}

\usepackage{url}
\usepackage{tabularx}
\usepackage{float}
\usepackage{listings}
\usepackage{tcolorbox}
\tcbuselibrary{skins, breakable}
\usepackage{chngcntr}

\definecolor{cvprblue}{rgb}{0.81,0.89,0.74}
\usepackage[pagebackref,breaklinks,colorlinks,allcolors=cvprblue]{hyperref}
\hypersetup{
    colorlinks=true,
    linkcolor=blue,
    urlcolor=blue
}
\def\paperID{*****} 
\def\confName{CVPR}
\def\confYear{2026}

\title{See\&Say: Vision Language Guided Safe Zone Detection for Autonomous Package Delivery Drones}

\author{
Mahyar Ghazanfari, Peng Wei \\
George Washington University \\
{\small \texttt{\{mahyar.ghazanfari, pwei\}@gwu.edu}}
}
\begin{document}
\maketitle

\begin{abstract}
Autonomous drone delivery systems are rapidly advancing, but ensuring safe and reliable package drop-offs remains highly challenging in cluttered urban and suburban environments where accurately identifying suitable package drop zones is critical. Existing approaches typically rely on either geometry-based analysis or semantic segmentation alone, but these methods lack the integrated semantic reasoning required for robust decision-making. To address this gap, we propose \textit{See\&Say}, a novel framework that combines geometric safety cues with semantic perception, guided by a Vision--Language Model (VLM) for iterative refinement. The system fuses monocular depth gradients with open-vocabulary detection masks to produce safety maps, while the VLM dynamically adjusts object category prompts and refines hazard detection across time, enabling reliable reasoning under dynamic conditions during the final delivery phase. When the primary drop-pad is occupied or unsafe, the proposed \textit{See\&Say} also identifies alternative candidate zones for package delivery. We curated a dataset of urban delivery scenarios with moving objects and human activities to evaluate the approach. Experimental results show that \textit{See\&Say} outperforms all baselines, achieving the highest accuracy and IoU for safety map prediction as well as superior performance in alternative drop zone evaluation across multiple thresholds. These findings highlight the promise of VLM-guided segmentation-depth fusion for advancing safe and practical drone-based package delivery. The codebase, dataset, and all supplementary materials, including the prompts and threshold settings used during inference, are available \href{https://github.com/Mahyar-GH79/See-Say-Vision-Language-Guided-Safe-Zone-Detection-for-Autonomous-Package-Delivery-Drones}{here}.
\end{abstract}

\section{Introduction}
Unmanned aerial vehicles (UAVs), commonly known as drones, have experienced rapid growth in popularity and are now integral to both research and industry \cite{tezza2019state}. They are increasingly employed across diverse civilian applications, including last-mile package delivery \cite{wing, zipline}. Autonomous delivery missions typically require drones to depart from a distribution center, navigate to a target location (customer's home), safely deploy a package, and return to base \cite{dos2023package}. Among these steps, the package drop phase is particularly challenging: designated delivery platform or pad may be obstructed, unsafe, or inaccessible at the time of drone's arrival. In such cases, drones must autonomously flag the primary drop-pad as unsafe and then identify and evaluate alternative safe zones. This challenge is further amplified in cities and suburbs, where dense obstacles, dynamic pedestrian activity, and limited open spaces create a highly complex perception and planning problem.

Prior works have primarily relied on geometry-based methods using depth information \cite{tan2025vislanding, vazquezlanding} or semantic approaches based on object detection and segmentation \cite{abdollahzadeh2022safe, de2025vision, kinahan2021image, song2022multi, vu2025transformer}. Other studies have addressed emergency landing in urban environments \cite{slama2023risk, moortgat2024autonomous, mitroudas2024light, sinhmar2024practical}. From a perception standpoint, these methods are effective in certain conditions but do not jointly exploit geometric and semantic information for robust safety assessment. Efforts that fuse multimodal inputs such as camera and LiDAR \cite{chen2020robust, liu2022real} improve spatial awareness but still lack the reasoning needed to generate explainable safety maps which is an essential factor for building trust and enabling reliable autonomy in safety-critical delivery tasks.

To address this gap, we propose \textit{See\&Say}, a novel framework that integrates monocular depth estimation, semantic segmentation, and a VLM for iterative refinement of safety perception. Inspired by the rapid progress of VLMs in UAV applications, we exploit their zero-shot generalization and reasoning capabilities as supervisory agents for both safety map generation and alternative drop zone selection.

Our framework operates solely on monocular RGB input, eliminating the need for LiDAR or other additional sensors. It follows a multi-stage pipeline that fuses geometric analysis with semantic understanding for comprehensive safety assessment. First, Depth-Anything V2 \cite{yang2024depth}, a state-of-the-art monocular depth estimator, generates depth-gradient maps to distinguish flat from non-flat regions. In parallel, DINO-X \cite{ren2024dinoxunifiedvisionmodel}, an open-vocabulary detector, provides prompt-based segmentation of hazardous objects. Fusing these geometric and semantic information produces an initial safety map. Since this preliminary map may miss relevant hazards, a VLM adaptively refines the DINO-X prompts based on scene context, improving segmentation quality and map accuracy. Beyond single-frame perception, the VLM analyzes temporal batches of frames to assess the safety of the primary drop-pad. When the primary drop-pad is deemed unsafe or obstructed, an auxiliary VLM agent ranks alternative drop zones according to human safety preferences.

Due to the absence of suitable datasets for this problem, we curated a custom dataset consisting of three residential front- and backyard videos featuring diverse objects, varying lighting conditions, and human activities. Experimental results show that \textit{See\&Say} consistently outperforms all baselines in primary drop-pad safety, safety map Intersection over Union (IoU), and alternative drop zone evaluation (Average Precision (AP) and Receiver Operating Characteristic (ROC)), establishing a strong foundation for VLM-guided safe package delivery.

\noindent In summary, our contributions are as follows:
\begin{itemize}[leftmargin=*,noitemsep,topsep=0pt]
    \item We propose \textit{See\&Say}, the first VLM-guided safe drop zone detection framework that introduces explainability and reasoning into safety decisions for package delivery drones.
    \item We design a lightweight geometric--semantic fusion pipeline, combining depth gradients with open-vocabulary segmentation outputs to generate comprehensive safety maps.
    \item We curate a dataset tailored for package delivery safety in cluttered suburb delivery scenes, and demonstrate significant improvements over baseline methods across multiple evaluation metrics.
\end{itemize}

\begin{figure*}[t]
    \centering
    \setlength{\fboxsep}{3pt}
    \setlength{\fboxrule}{0.8pt}

    \fcolorbox{black}{white}{%
        \includegraphics[width=\textwidth]{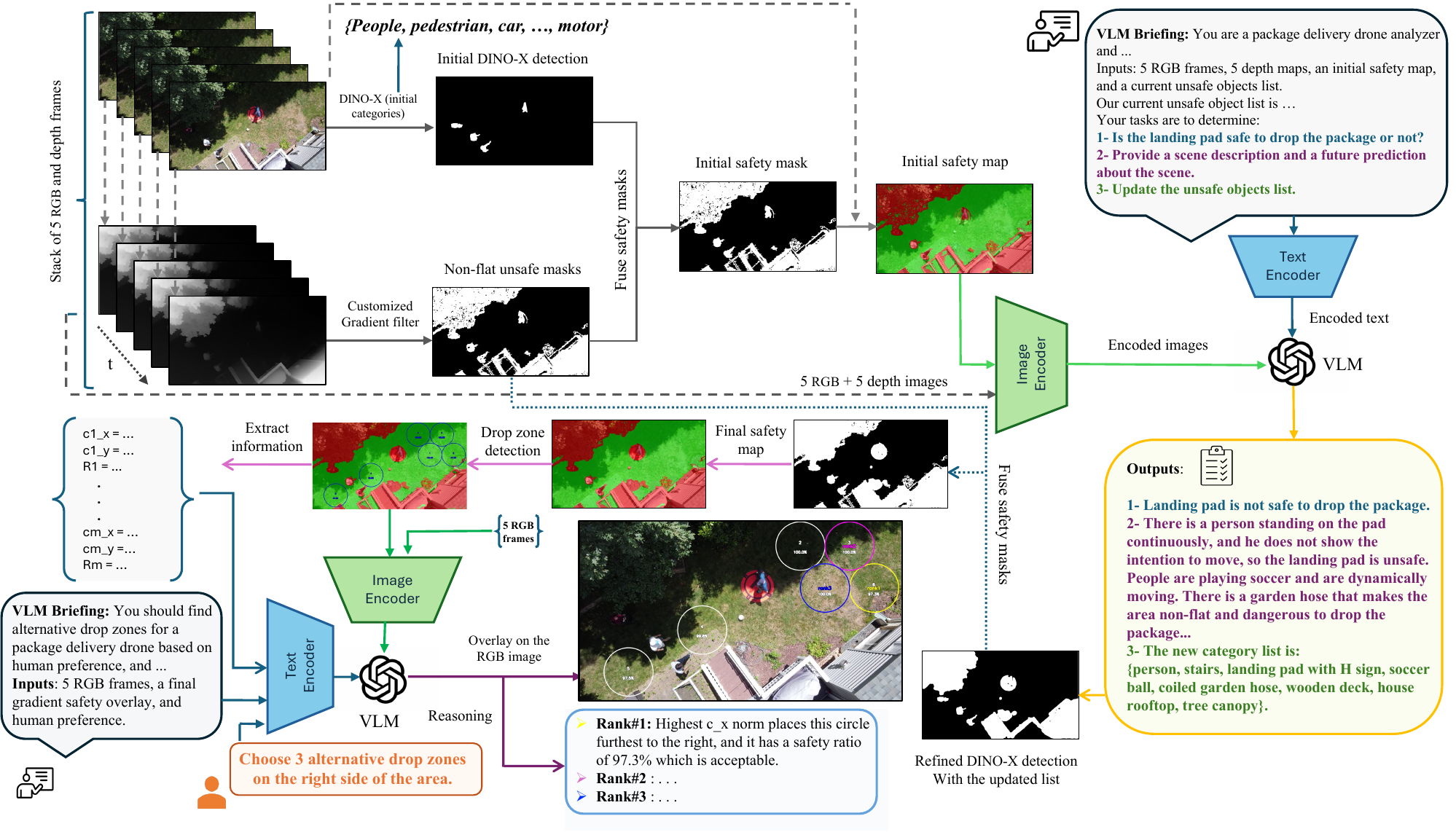}%
    }

    \caption{Overview of the proposed \textit{See\&Say} framework. The system takes batches of five RGB frames and corresponding monocular depth maps as input. Depth gradients provide geometric cues for flatness and obstacle detection, while DINO-X produces open-vocabulary semantic hazard masks. These initial maps are fused to form a preliminary safety overlay. A VLM refines detection prompts using temporal RGB--depth context, improving semantic segmentation and updating the safety map. A second VLM agent evaluates the primary drop pad's safety over time and ranks alternative candidate drop zones according to user-defined preferences. The final outputs include refined safety maps, primary drop-pad safety decisions, scene description, and a ranked list of safe alternative zones.}

    \label{fig:1}
\end{figure*}

\section{Related Work}

\subsection*{\textbf{Geometry-Based Safe Zone Detection}}
Geometry-based methods rely primarily on depth information to identify flat terrain for safe landing operations. \cite{vazquezlanding} employed transformer architectures to classify landing zones directly from depth maps, while SafeUAV~\cite{marcu2018safeuav} developed a lightweight CNN trained on synthetic data to jointly predict depth and obstacle classes. VisLanding~\cite{tan2025vislanding} advanced this direction by refining monocular predictions using Dense Prediction Transformers~\cite{ranftl2021vision}. These methods achieve reliable flatness estimation and terrain suitability assessment; however, they overlook semantic hazards, thereby limiting robustness in dynamic environments.

\subsection*{\textbf{Semantic and Detection-Based Methods}}
Semantic approaches rely on object detection and segmentation to construct safety maps. \cite{abdollahzadeh2022safe} employed U-Nets to regress continuous pixel-level safety values, while \cite{de2025vision} developed networks that assign dynamic risk scores under challenging conditions such as moving vehicles, pedestrians, or varying illumination. \cite{kinahan2021image} classified safe versus unsafe regions directly from RGB input, providing interpretable binary maps for real-time use. These methods significantly advance semantic hazard recognition and contextual awareness, but they do not reason about geometric flatness and remain constrained to fixed categories, which restricts generalization to novel hazards and ultimately reduces explainability and trustworthiness in safety-critical UAV operations like package delivery.

\subsection*{\textbf{Sensor Fusion for UAV Perception}}
RGB imagery provides rich semantic cues but is often insufficient for safety-critical UAV tasks. To address this limitation, several studies fuse LiDAR-derived 3D point clouds with 2D imagery to enhance spatial perception for landing and navigation \cite{bultmann2021real,chen2020robust}. These approaches improve both geometric and semantic awareness, yet they still lack temporal reasoning and explainability, which are critical for reliable safety assessments. Moreover, LiDAR remains a costly and power-demanding sensor, making it impractical for lightweight and resource-constrained package delivery UAVs.

\subsection*{\textbf{Vision--Language Models in UAVs}}
Recent works apply VLMs to UAVs, supporting tasks from mission planning~\cite{tabrizian2025chain} to perception and control~\cite{tian2025uavs}. Examples include using LLaVA-1.5 for weather analysis from UAV imagery~\cite{kim2024weather} and integrating VLMs for visibility estimation~\cite{sakaino2023dynamic}. However, these efforts focus on environmental understanding rather than safety-critical decision-making. To our knowledge, no prior work applies VLMs to adaptive safety perception. Our framework, addresses this gap by producing explainable, scene-aware evaluations of the environment.

\section{Methodology}

This section presents the \textit{See\&Say} framework for autonomous safe drop zone identification in UAV delivery scenarios. Figure~\ref{fig:1} illustrates the overall system architecture. The pipeline operates on batches of five RGB frames and their corresponding depth maps, fusing geometric cues from depth-gradient analysis with semantic hazard information from an open-vocabulary detector. Two VLM agents provide reasoning: (\emph{i}) one refines object detector prompts using the five-frame context, and (\emph{ii}) the other ranks candidate drop zones conditioned on human safety preferences. We use the GPT-o3 reasoning model for both agents. The individual components are detailed in the following subsections together with their mathematical notation.

We process the stream in fixed windows (batches) of five frames
$\mathcal{B}_k=\{I_{t-4},I_{t-3},I_{t-2},I_{t-1},I_t\}$.
Only the fifth frame $I_t$ (``decision frame'') is scored for dropping the package,
while the preceding four frames provide spatial--temporal context to depth and
vision--language refinement.

\noindent\textbf{Depth Estimation.}
For each $I_\tau\in\mathcal{B}_k$, we predict dense depth with Depth-Anything~V2:
\begin{equation}
    D_\tau=\mathcal{M}_\theta(I_\tau)\in\mathbb{R}^{H\times W}.
\end{equation}
where $I_\tau$ is the $\tau$-th frame from the batch and $\mathcal{M}_\theta$ denotes the monocular depth estimator. To remove per-frame scale, we normalize
\begin{equation}
    \hat{D}_\tau=\frac{D_\tau-\min(D_\tau)}{\max(D_\tau)-\min(D_\tau)+\epsilon},\quad \hat{D}_\tau\in[0,1],
\end{equation}
with $\epsilon>0$ for numerical stability.

\noindent\textbf{Gradient-Based Flatness (Depth Geometry).}
Geometric hazards are detected through gradient analysis of smoothed depth estimates:
\begin{align}
    D^{\text{smooth}}_\tau &= G_\sigma * \hat{D}_\tau, \\
    \|\nabla D_\tau\|(x,y) &= \sqrt{\left(\partial_x D^{\text{smooth}}_\tau\right)^2+\left(\partial_y D^{\text{smooth}}_\tau\right)^2}.
\end{align}
Where $G_\sigma$ is a gaussian smoothing filter. A pixel is \emph{flat} if a $w\times w$ window $\mathcal{W}_{x,y}$ around it contains a sufficient fraction of
sub-threshold gradients:
\begin{equation}
    \frac{1}{|\mathcal{W}_{x,y}|}\sum_{(u,v)\in\mathcal{W}_{x,y}}\mathds{1}\!\left(\|\nabla D_\tau\|(u,v)<\tau_g\right)\ge \gamma,
\end{equation}
with gradient threshold $\tau_g$ and flatness ratio $\gamma$.
After morphological open/close and small-component suppression, we obtain a binary mask
$M^{\text{flat}}_\tau\in\{0,1\}^{H\times W}$ (\(1\)=flat/safe).
Its complement yields a geometric \emph{unsafe} mask
$U^{\text{grad}}_\tau = \mathbf{1}-M^{\text{flat}}_\tau$ (\(1\)=unsafe).

\noindent\textbf{DINO-X Open-Vocabulary Hazard Detection.}
Semantic hazards are detected through DINO-X segmentation guided by a prompt vocabulary $\mathcal{P}$ containing initial default classes that is chosen by the user
(e.g., \textit{person}, \textit{car}, \textit{bus}, \dots).
Given $\mathcal{P}$, DINO-X produces per-class spatial masks $\{M_c\}_{c\in\mathcal{P}}$.
We aggregate them into an \emph{unsafe map}:
\begin{equation}
    U^{\text{dinox}}_t(x,y) = \max_{c\in\mathcal{P}} M_c(x,y),
\end{equation}
and binarize at $\theta_d$ to obtain
\[
B^{\text{dinox}}_t = \mathbf{1}\{U^{\text{dinox}}_t \ge \theta_d\}.
\]

\noindent\textbf{VLM-in-the-Loop Prompt Refinement.}
Let $\mathcal{P}^{(0)}$ denote the initial class (prompt) list.
From the first pass, we construct an \emph{initial safety overlay} on frame $I_t$ by fusing the semantic unsafe map $B^{\text{dinox}}_t$ with the gradient-unsafe mask $U^{\text{grad}}_t$:
\[
\text{overlay}_t = B^{\text{dinox}}_t \lor U^{\text{grad}}_t.
\]

\begin{figure}[t]
    \setlength{\fboxsep}{1.5pt}
    \setlength{\fboxrule}{0.8pt}

    \fcolorbox{black}{white}{%
        \includegraphics[width=\columnwidth]{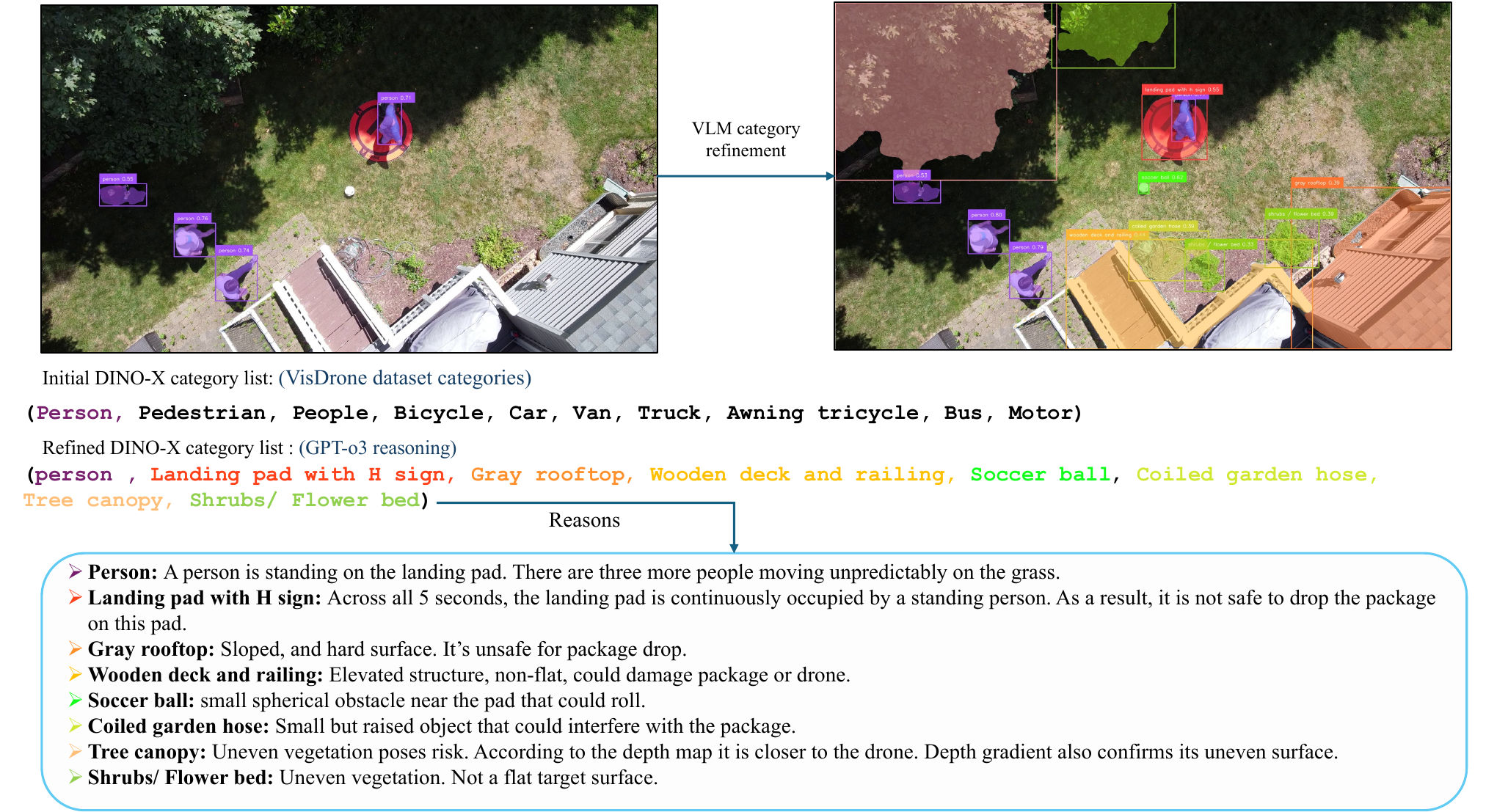}%
    }

\caption{The VLM refines the object categories used by DINO-X based on the actual scene content, improving hazard detection and safety map accuracy.}

    \label{fig:2}
\end{figure}

Given the batch of five RGB frames $\{I_\tau\}_{\tau=t-4}^t$, their normalized depth maps $\{\hat{D}_\tau\}_{\tau=t-4}^t$, and the initial safety overlay $\text{overlay}_t$, a VLM produces a refined prompt list via
\begin{equation}
    \mathcal{P}^{(k+1)} =
    \mathrm{Refine}\!\left(
        \mathcal{P}^{(k)},
        \{I_\tau\}_{\tau=t-4}^{t},
        \{\hat{D}_\tau\}_{\tau=t-4}^{t},
        \text{overlay}_t
    \right),
\end{equation}
where $\mathrm{Refine}(\cdot)$ is the VLM's instruction-following operator.
It updates the class list by (\emph{i}) adding/removing unsafe objects, (\emph{ii}) merging synonyms, and (\emph{iii}) adjusting specificity (e.g., \textit{vehicle} $\mapsto$ \{\textit{car}, \textit{truck}, \textit{bus}\}). Figure \ref{fig:2} shows refinement of the category list using the VLM.
In addition, the VLM returns an auxiliary output tuple
\[
\big(\text{safe\_flag}_t,~ \text{scene\_desc}_t,~ \text{future\_pred}_t\big),
\]
where
(\emph{i}) $\text{safe\_flag}_t \in \{0,1\}$ indicates whether dropping the package at the primary drop-pad is safe,
(\emph{ii}) $\text{scene\_desc}_t$ is a short textual summary of RGB+depth context, and
(\emph{iii}) $\text{future\_pred}_t$ forecasts near-future scene evolution.

Finally, we re-run DINO-X on $I_t$ with $\mathcal{P}^{(k+1)}$ to obtain the \emph{refined} semantic unsafe map $B^{\text{dinox,ref}}_t$, which is used in the final fusion step.

\noindent\textbf{Safety Map Fusion.}
Each module yields an \emph{unsafe} map on the decision frame $I_t$.
For ablations we use a logical union (pessimistic):
\begin{equation}
    B^{\text{final}}_t = B^{\text{dinox,ref}}_t \,\lor\, U^{\text{grad}}_t,
\end{equation}
with \(\lor\) applied pixelwise.\\ Figure \ref{fig:qual_grid_3x7} shows the iterative process of finding the final safety map after multiple steps of refinement.

\begin{figure}[t]
    \setlength{\fboxsep}{3pt}
    \setlength{\fboxrule}{0.8pt}

    \fcolorbox{black}{white}{%
        \includegraphics[width=0.97\columnwidth]{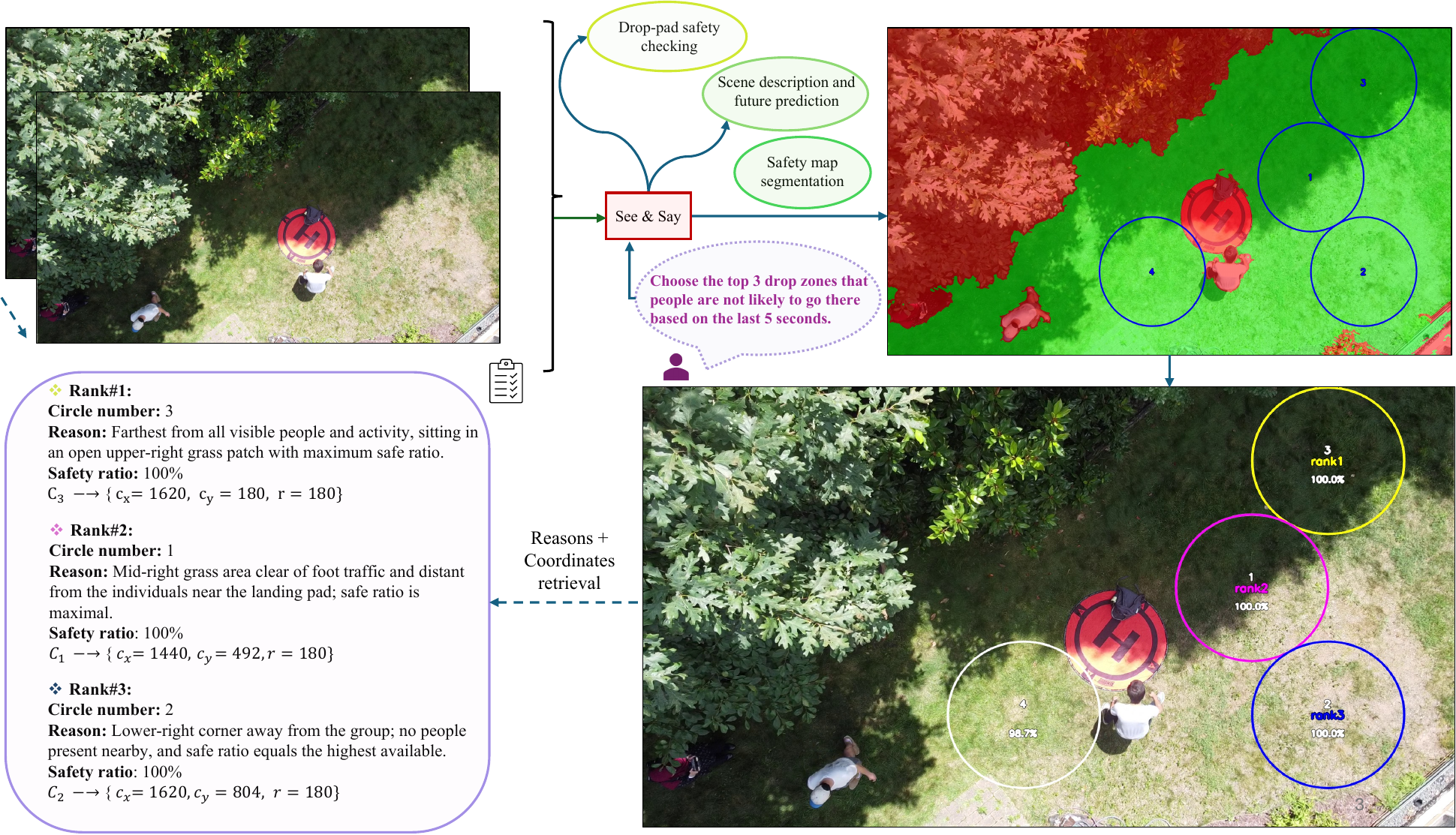}%
    }

   \caption{The second VLM agent in \textit{See\&Say} incorporates human preferences into package drop zone selection, operating after the initial safety map is generated by the first VLM agent.}

    \label{fig:3}
\end{figure}

\noindent\textbf{Alternative Drop Zone Candidates on a Hex Grid.}
We generate circle candidates of radius \(r\) using a hexagonal lattice to cover the footprint:
\begin{equation}
    \mathcal{C}=\{c_i=(x_i,y_i,r)\},\quad \Delta x = 2r,\ \Delta y = \sqrt{3}\,r.
\end{equation}
When an ``H'' landing pad is detected in $I_t$, we set
\begin{equation}
    r=\tfrac{1}{2}\sqrt{w^2+h^2},
\end{equation}
where \(w,h\) are the bounding-box sides; otherwise we use a default \(r\).
For each candidate \(c\), let \(S_c\) be its disk support and \(B^{\text{final}}_t\) the final unsafe mask.
We define the \emph{safe ratio}
\begin{equation}
    \mathrm{safe}(c)=1-\frac{1}{|S_c|}\sum_{(x,y)\in S_c} B^{\text{final}}_t(x,y).
\end{equation}
Feasible circles satisfy \(\mathrm{safe}(c)\ge \eta\) (e.g., \(\eta=0.95\)).
With the help of the second VLM agent, we score drop zones by safety and human preference to choose the alternative package drop zones. Figure \ref{fig:3} demonstrates the process of choosing final drop zones based on human preference.

\section{Experiments}
In this part, we explain our dataset, baselines, evaluation metrics, and final results of our work.

\subsection{Dataset}
Existing UAV datasets such as ICG \cite{DroneDataset}, which contain drone imagery of urban environments, or MidAir \cite{fonder2019mid}, which focuses on natural scenes, do not fully capture the requirements of our scenario. Specifically, they lack settings tailored to package delivery drones that simultaneously involve landing zones, cluttered urban conditions, and multi-frame sequences for temporal analysis. To address this gap, we recorded three real-world scenes in suburban areas of Fairfax, VA, USA, including both front and back yards. These scenes featured diverse objects, varying lighting conditions (e.g., shadow and direct sunlight), and dynamic movements of people and objects. From the recordings, we selected 120 consecutive frames to maximize diversity, then organized them into batches of five frames each, which serve as the input to the \textit{See\&Say} framework. This resulted in 24 batches. All batches were segmented and annotated by a human annotator to ensure accurate safety maps of the drop zones. We used the curated dataset to evaluate the framework's performance on primary drop-pad safety assessment, environmental safety-map generation, and alternative drop zone selection.

\subsection{Baselines}

For the detection-based baseline, we employ YOLOv8 \cite{yolov8_ultralytics}, an enhanced version of the original YOLO algorithm \cite{redmon2016you}. As another detection model, we utilize RT-DETR \cite{zhao2024detrs}, a state-of-the-art transformer-based object detector. Both models are pretrained on the COCO dataset \cite{lin2014microsoft} and subsequently fine-tuned on the VisDrone dataset \cite{cao2021visdrone}, an aerial imagery benchmark with 10 object categories ranging from pedestrians to buses and motorcycles. To further improve results, we combine the bounding boxes predicted by these detectors with a segmentation model. Specifically, we employ SAM2 \cite{ravi2024sam2} to segment regions within the detected bounding boxes, after which the segmented outputs are converted into safety maps.

As an alternative segmentation-based approach, we evaluate DINO-X. We test this algorithm under two configurations: (\emph{i}) using the full VisDrone category list, where all detected objects are treated as hazards and only the remaining area is considered safe, and (\emph{ii}) using the \textit{``flat\_ground''} category, where the model explicitly searches for flat regions. In the latter case, the segmented flat area is labeled as safe, while all other regions are labeled unsafe.

Another baseline is the Depth-Based Flatness algorithm, denoted as \textit{Gradient} in Tables \ref{tab:merged-metrics} and \ref{tab:landing_candidates} and Figure \ref{fig:roc_pr_combined}. To ensure fair comparison, we further augment all detection and segmentation-based pipelines with the depth-gradient filter described earlier, thereby incorporating geometric information from the environment to enhance performance. In the case of DINO-X, we utilize the \textit{``flat\_ground''} category and enhance it by incorporating the depth-gradient filter. Finally, we evaluate a vanilla VLM (GPT-o3 reasoning) for safe zone detection. Since VLMs cannot directly output pixel-level safety masks, we prompt the model to predict bounding box coordinates of unsafe zones. These bounding boxes are then overlaid on the original image to generate a binary safe--unsafe map.


\begin{table}[t]
\centering
\caption{Success rate and reasoning status of different methods and frame settings.}
\label{tab:acc_reason_single}
\setlength{\tabcolsep}{4pt}
\renewcommand{\arraystretch}{1.2}
\resizebox{\columnwidth}{!}{%
\begin{tabular}{lcc}
\toprule
\textbf{Method} & \textbf{Success rate} & \textbf{Reasoning} \\
\midrule
1 - DINO-X (VisDrone\_category) & 0.483 & $\times$ \\
2 - DINO-X (Complete\_category) & 0.916 & $\times$ \\
3 - See \& Say (Ours, Single Frame) & 0.958 & \checkmark \\
4 - See \& Say (Ours, 5 Frames) & 0.975 & \checkmark \\
\bottomrule
\end{tabular}%
}
\end{table}

\begin{figure*}[t]
\centering
\setlength{\tabcolsep}{3pt}
\renewcommand{\arraystretch}{1}

\newcommand{\img}[1]{\includegraphics[width=0.132\textwidth]{#1}}

\begin{tabular}{*{7}{c}}
\scriptsize \textbf{RGB} &
\scriptsize \textbf{Monocular Depth} &
\scriptsize \textbf{Initial DINOx} &
\scriptsize \textbf{Depth Gradient} &
\scriptsize \textbf{Initial Combined} &
\scriptsize \textbf{Refined DINOx} &
\scriptsize \textbf{Final Combined} \\

\scriptsize (a) & \scriptsize (b) & \scriptsize (c) & \scriptsize (d) & \scriptsize (e) & \scriptsize (f) & \scriptsize (g) \\

\img{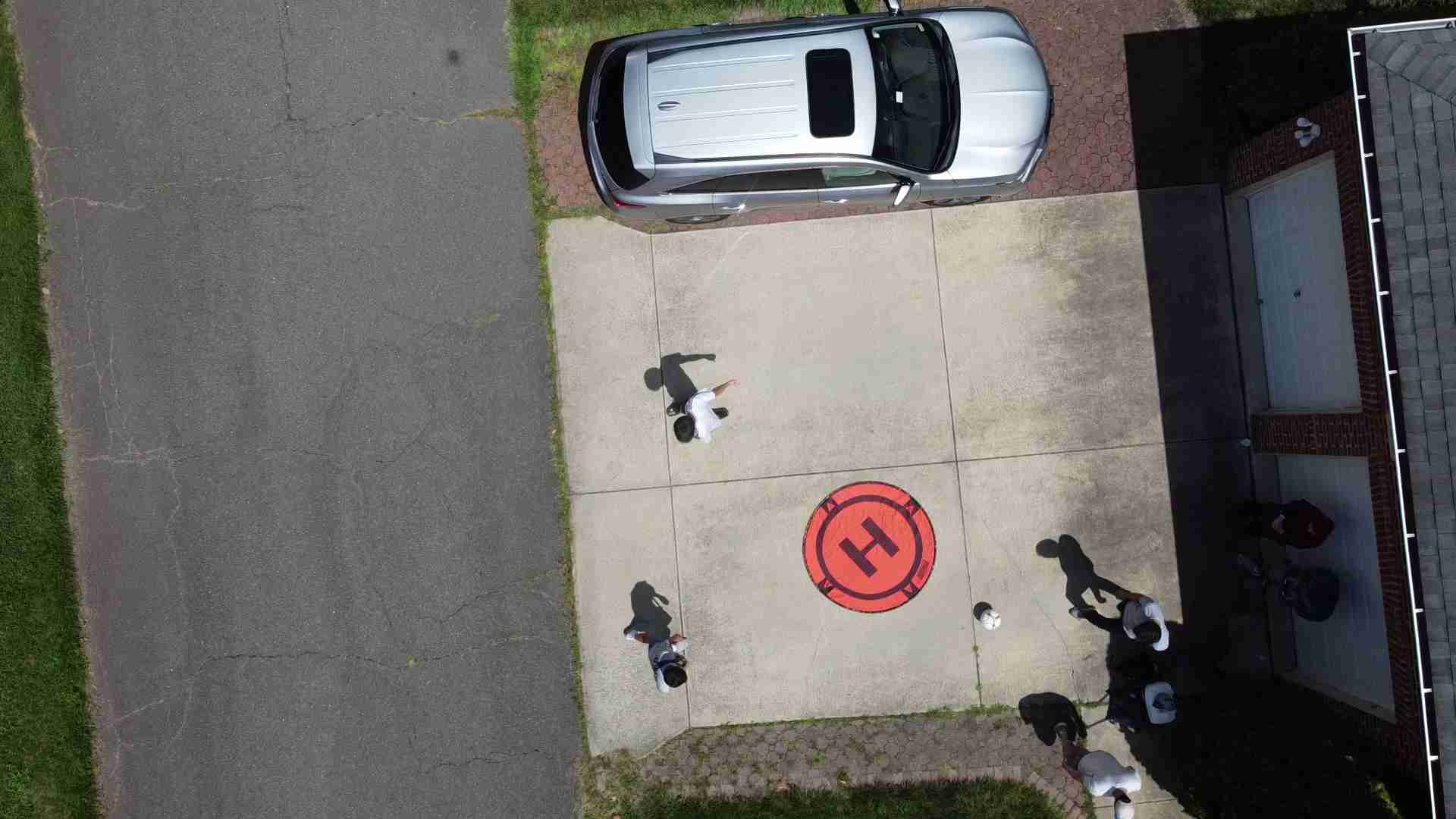} &
\img{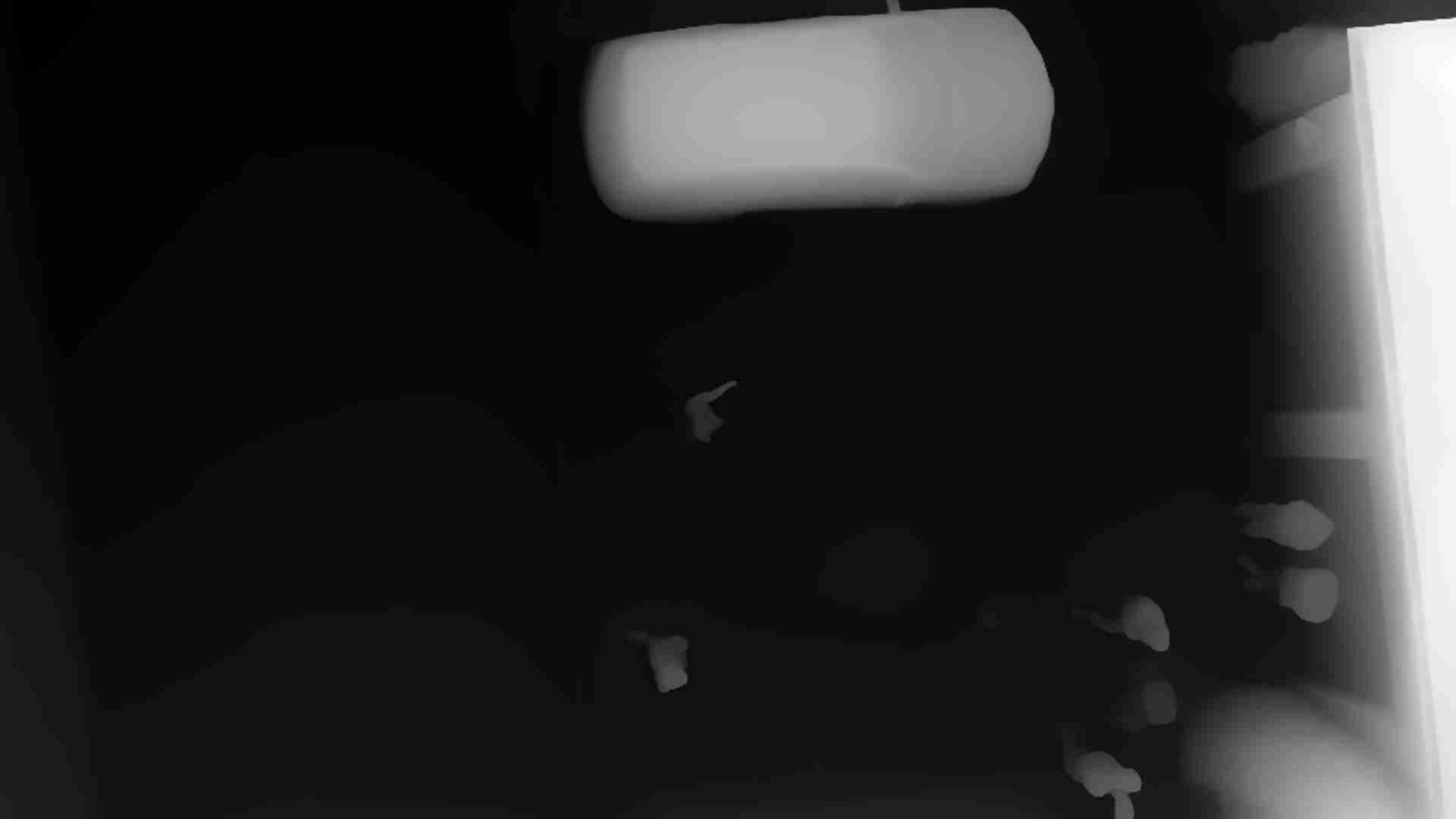} &
\img{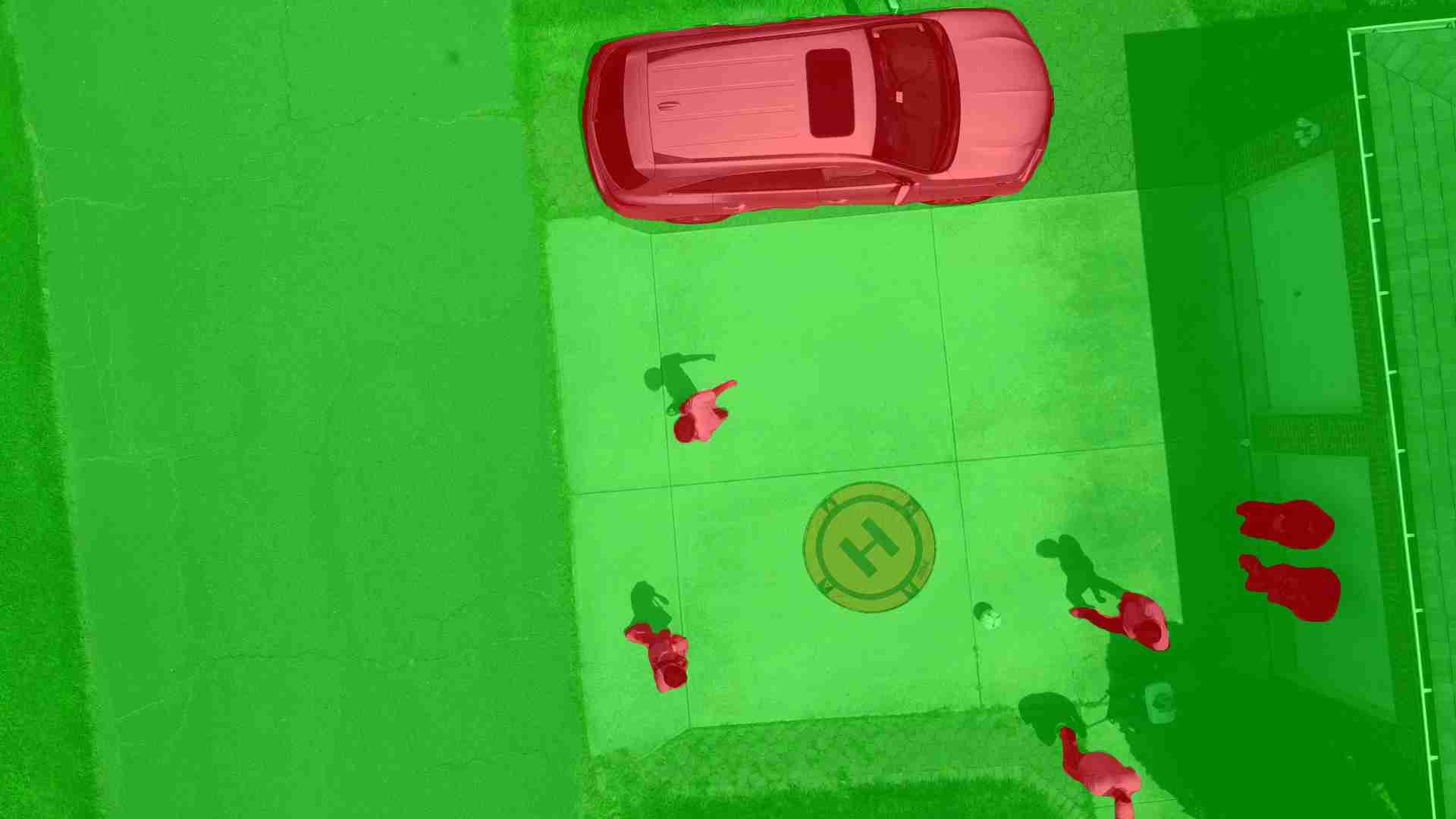} &
\img{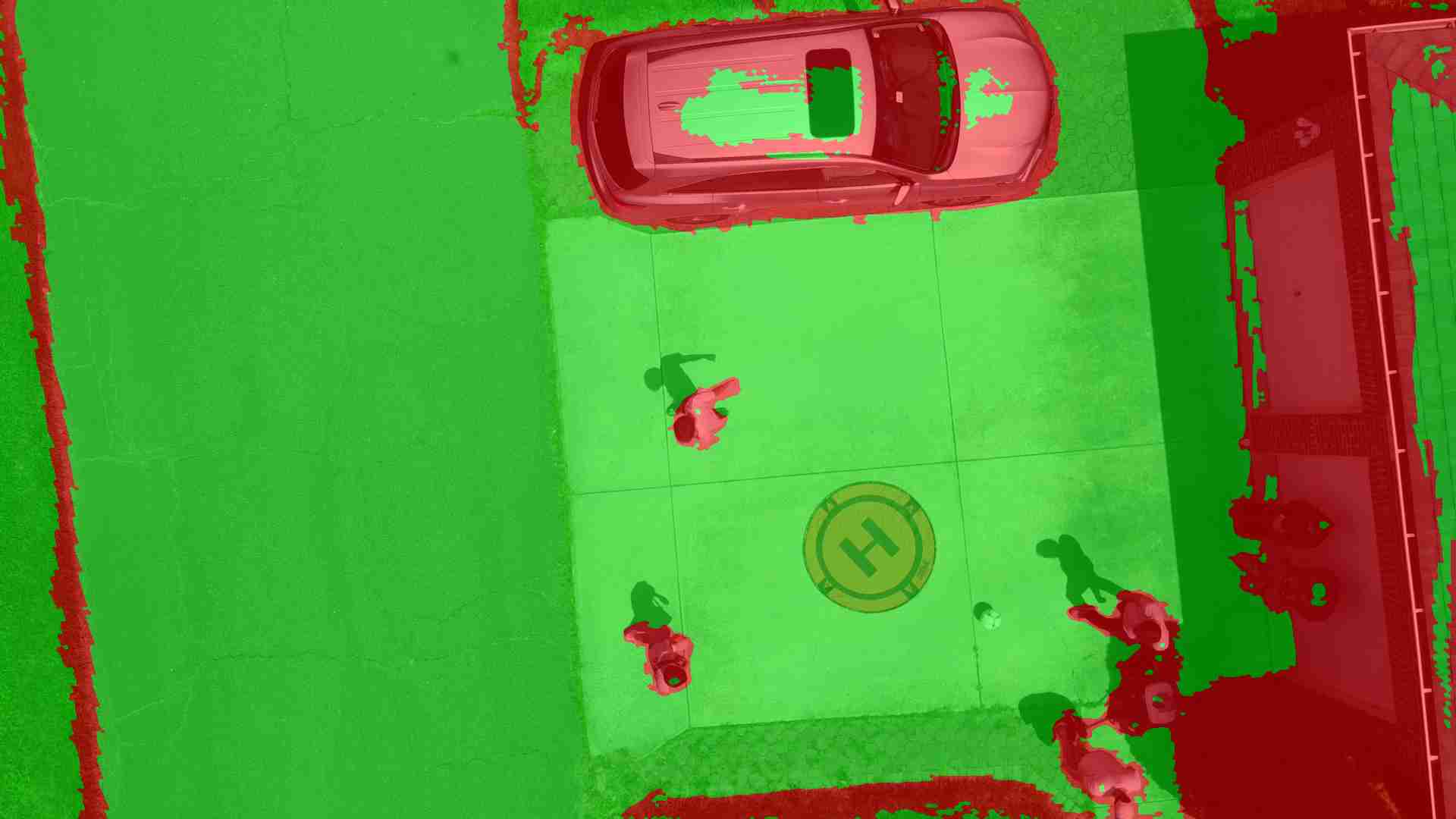} &
\img{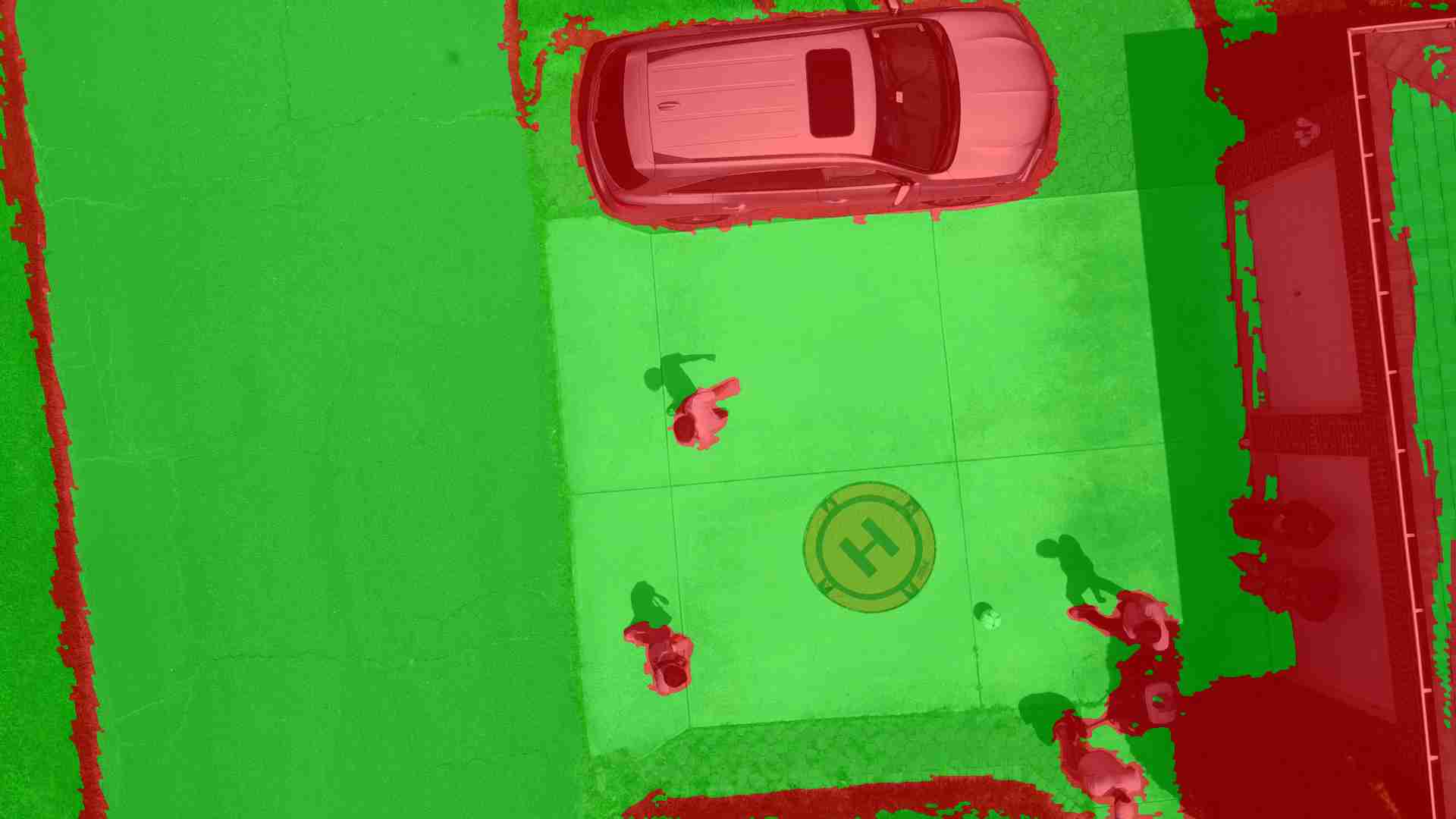} &
\img{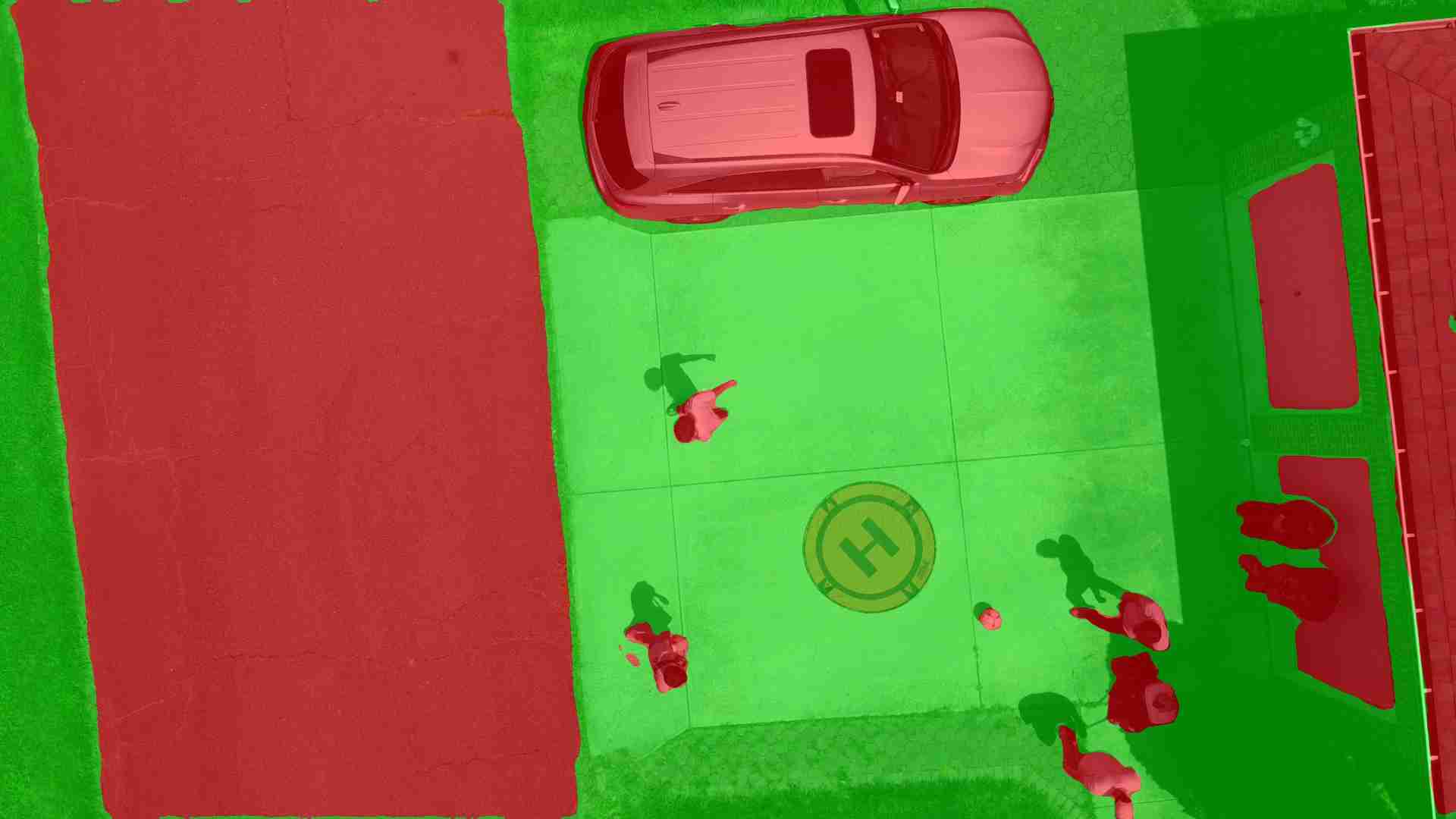} &
\img{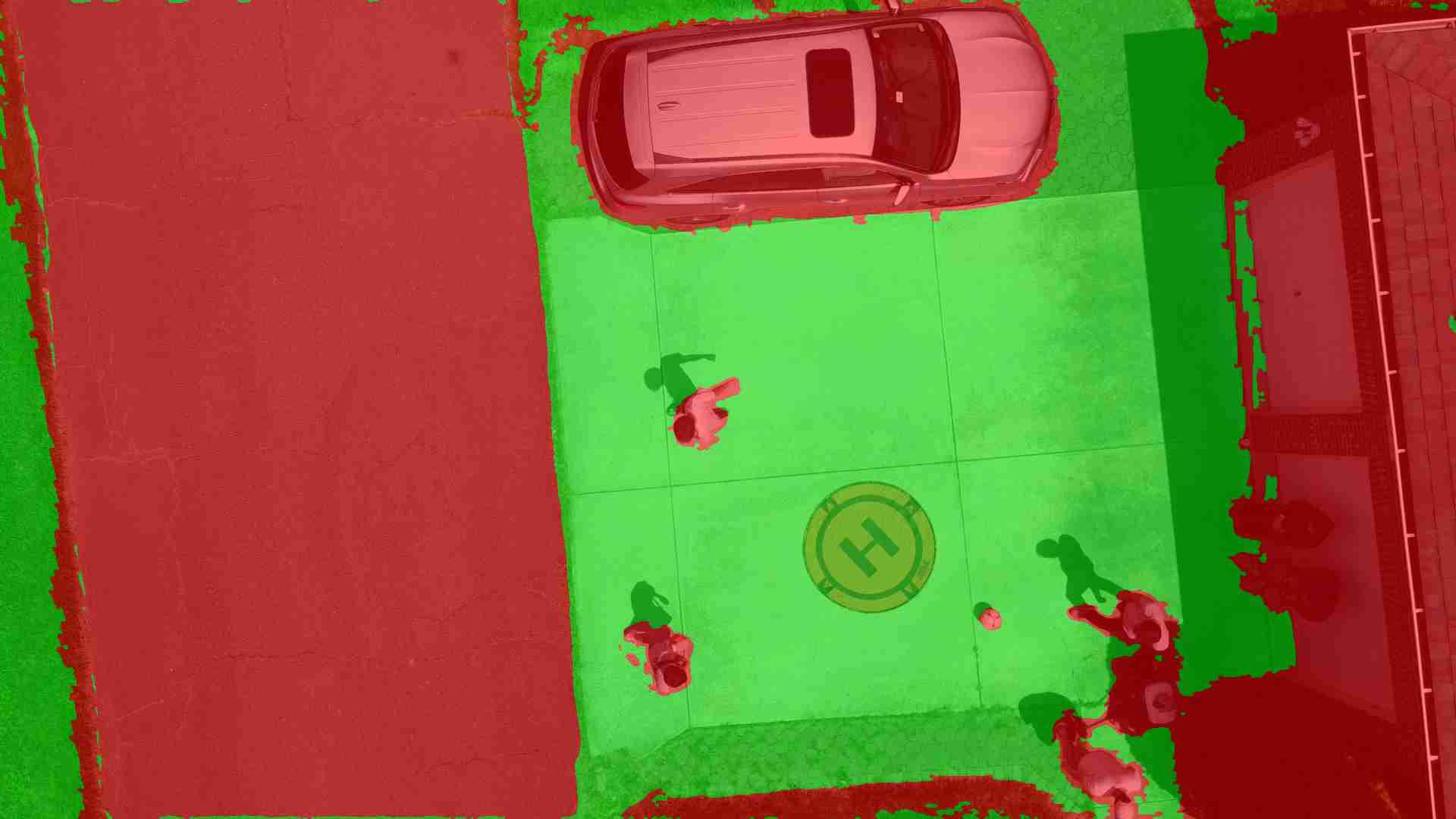} \\

\img{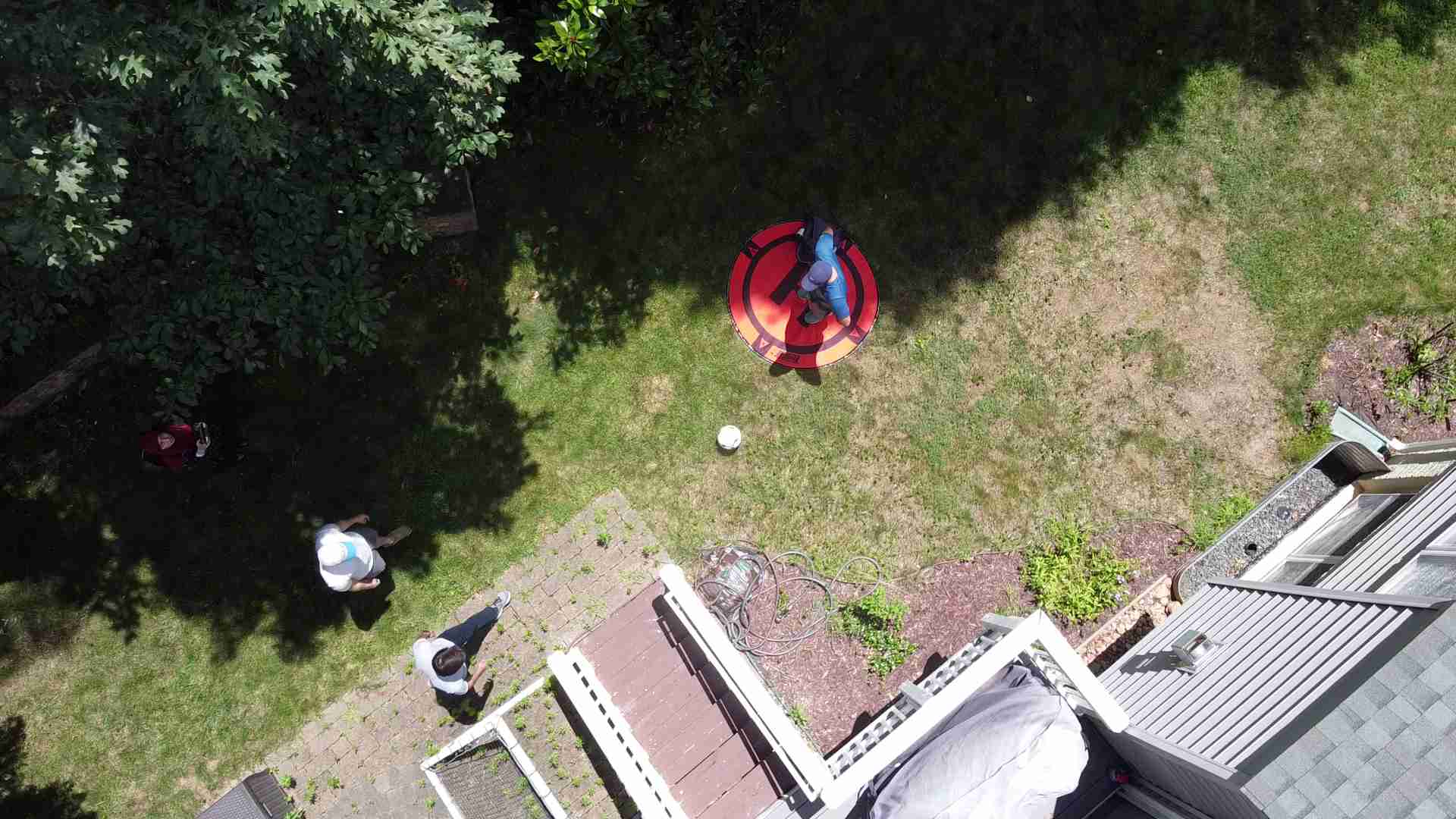} &
\img{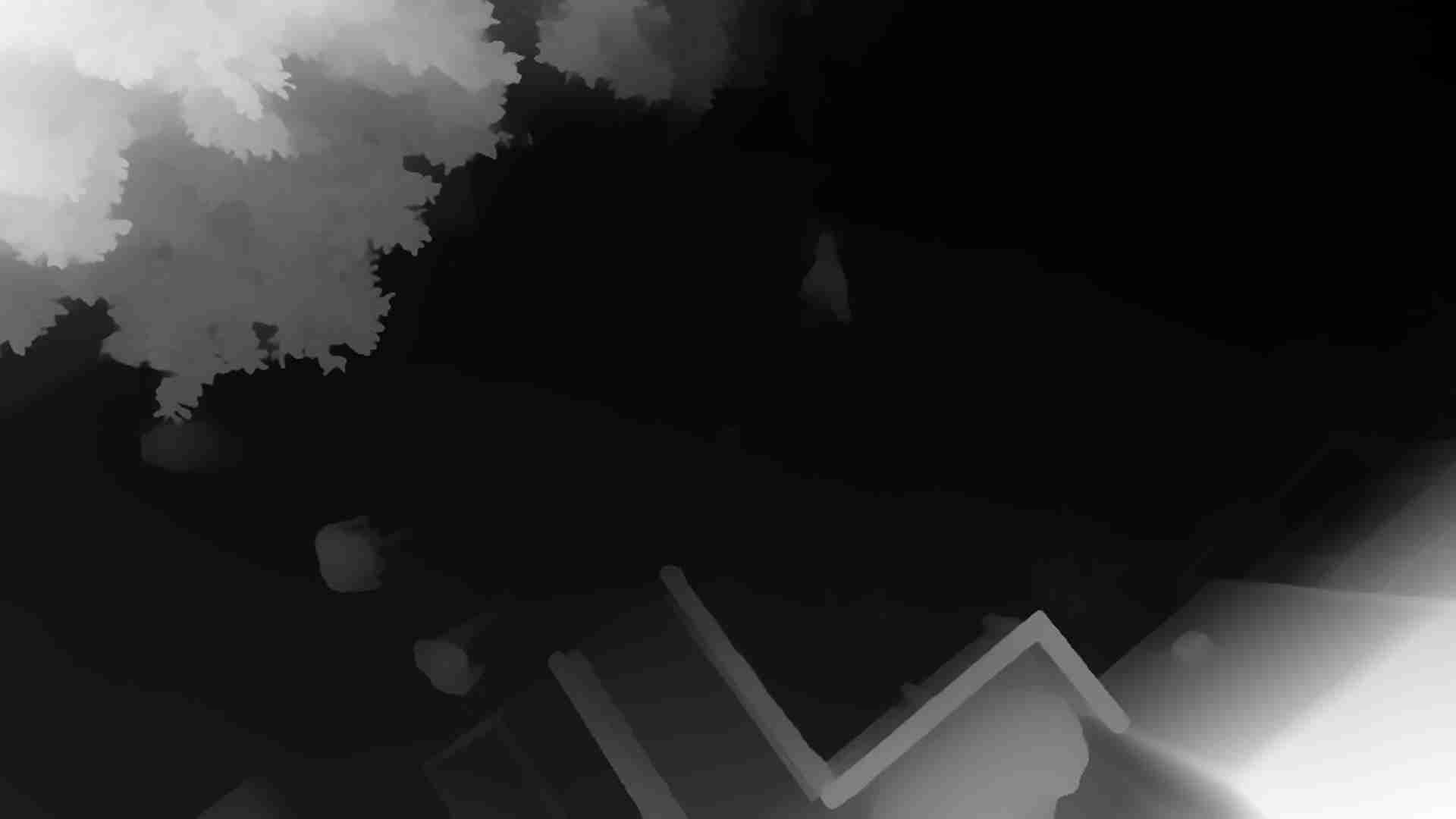} &
\img{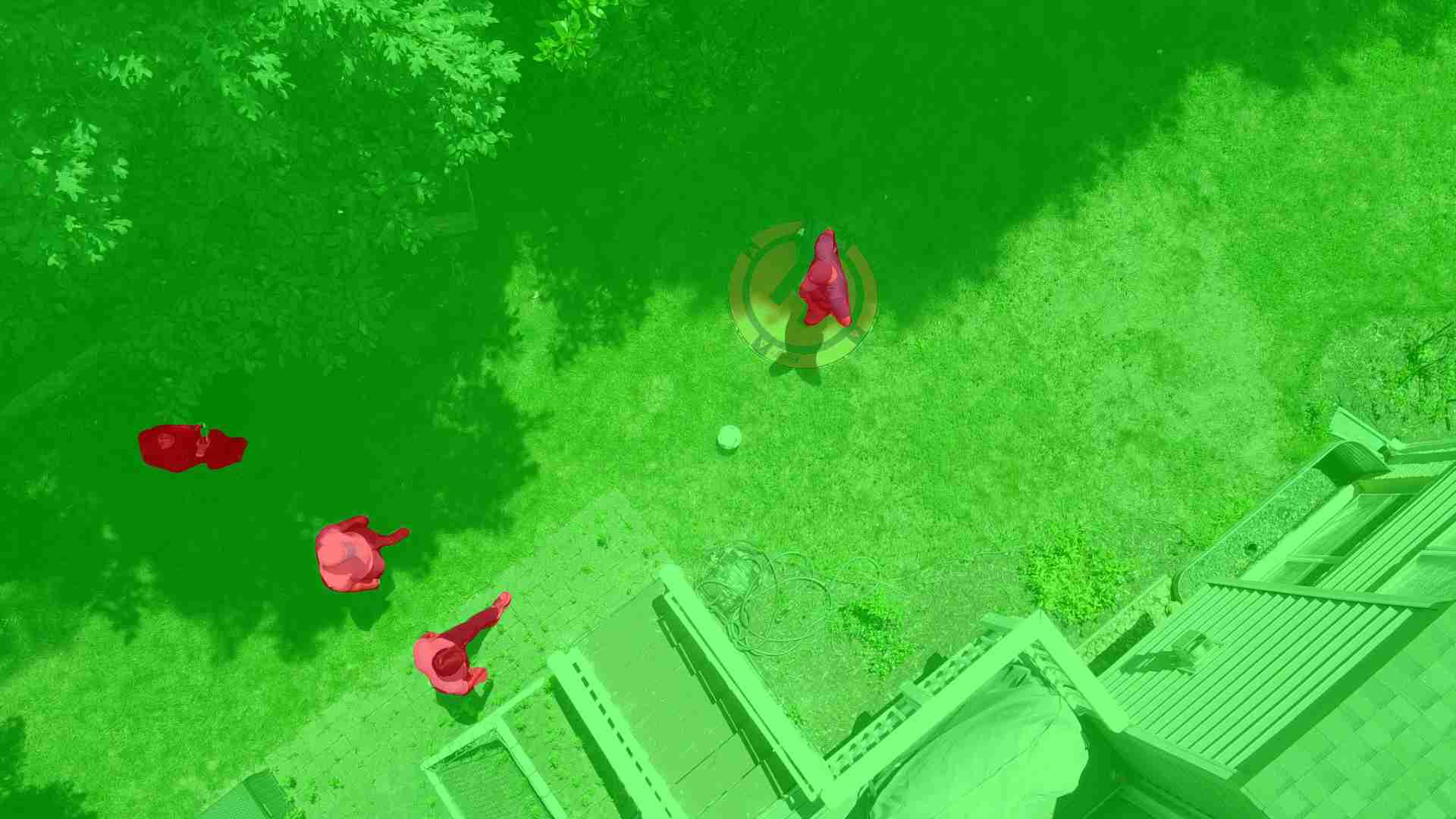} &
\img{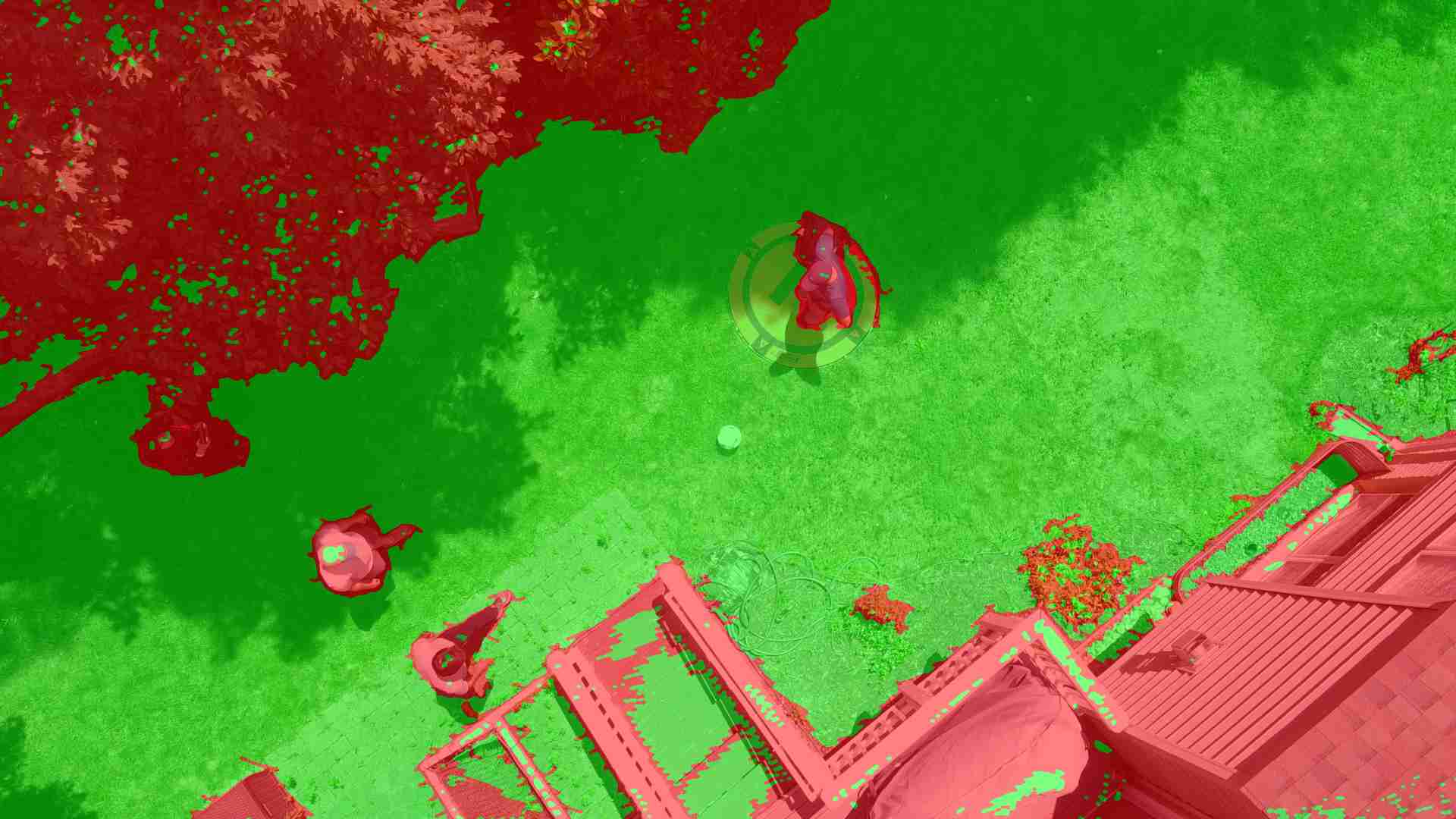} &
\img{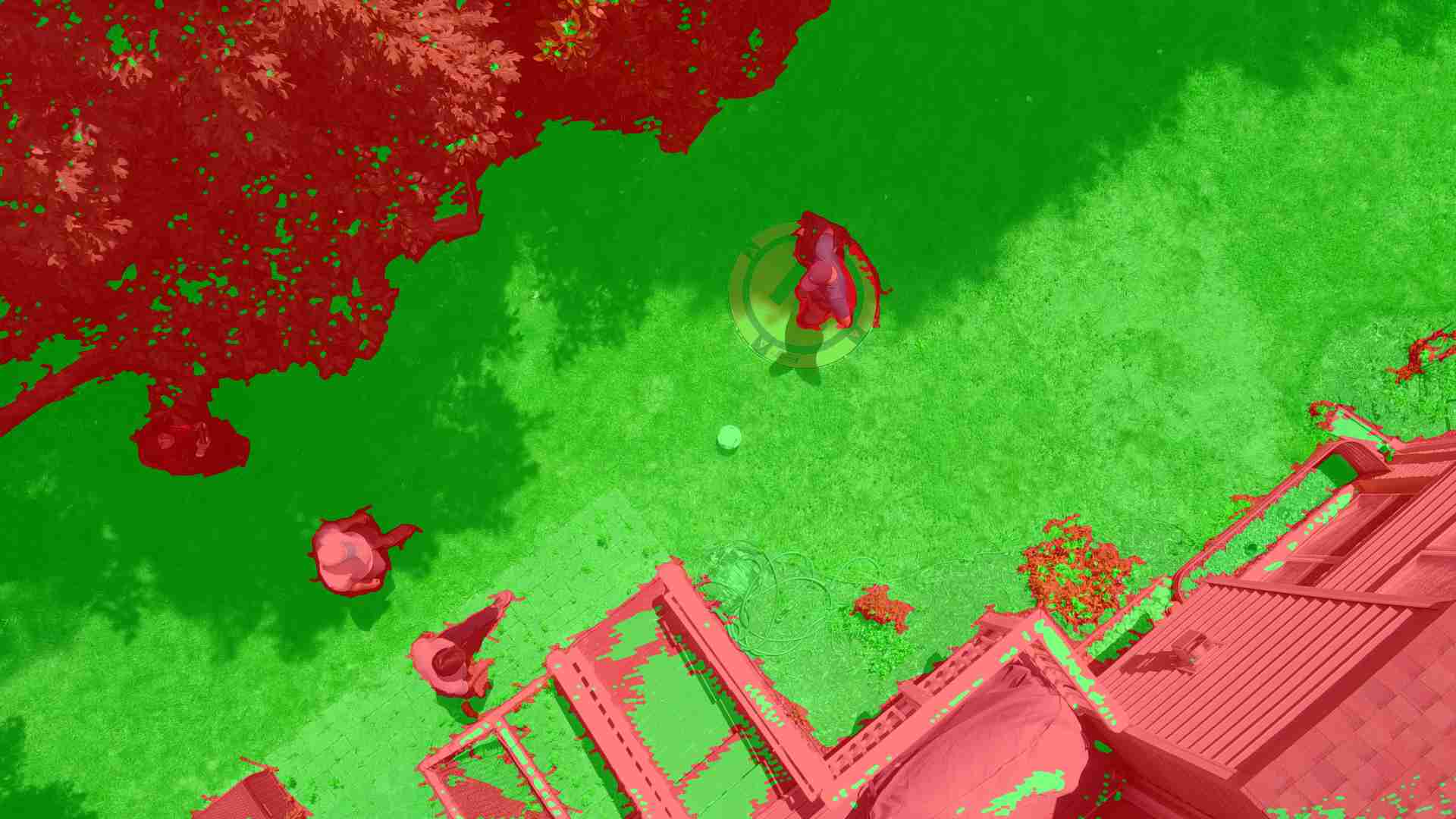} &
\img{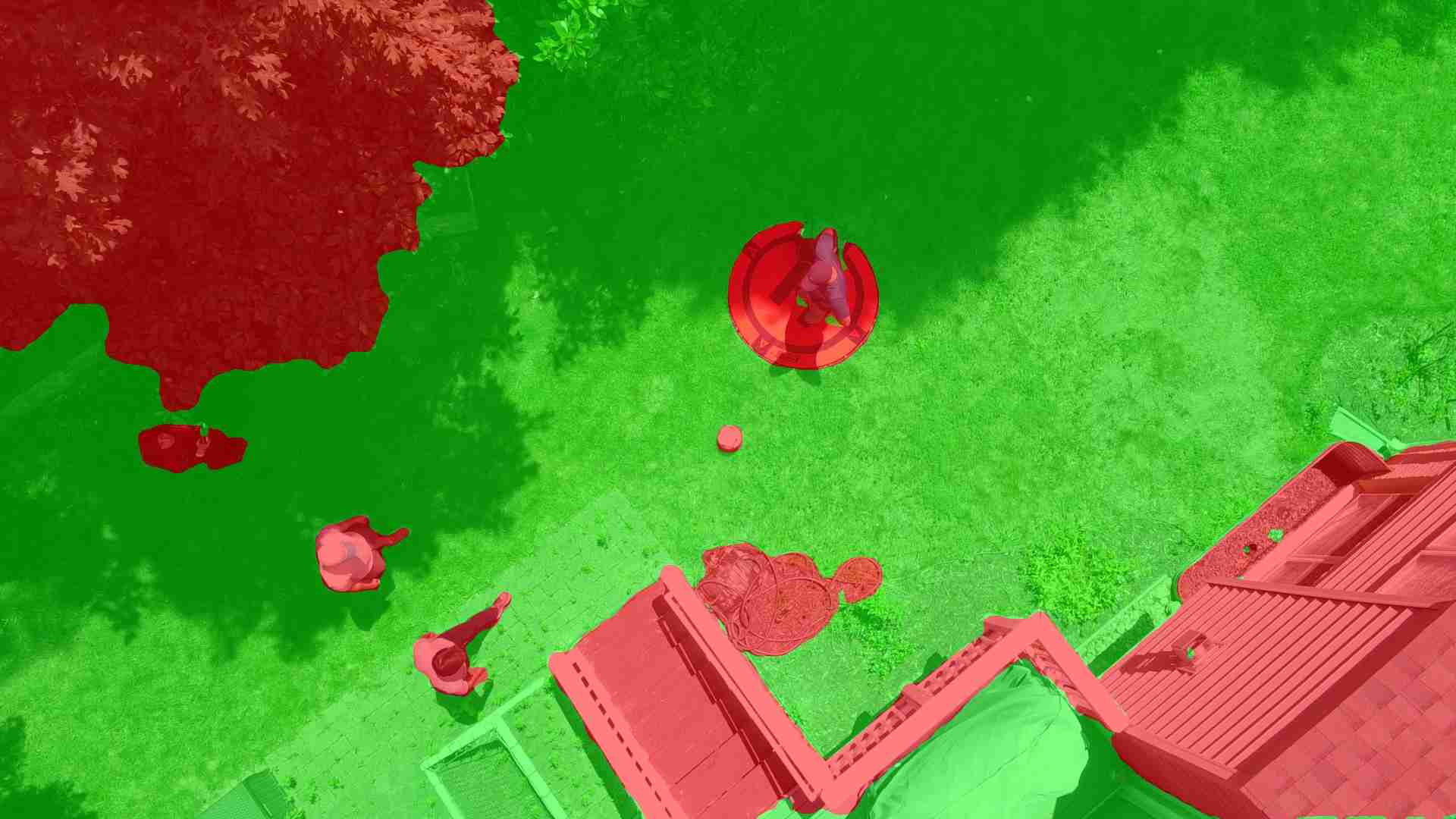} &
\img{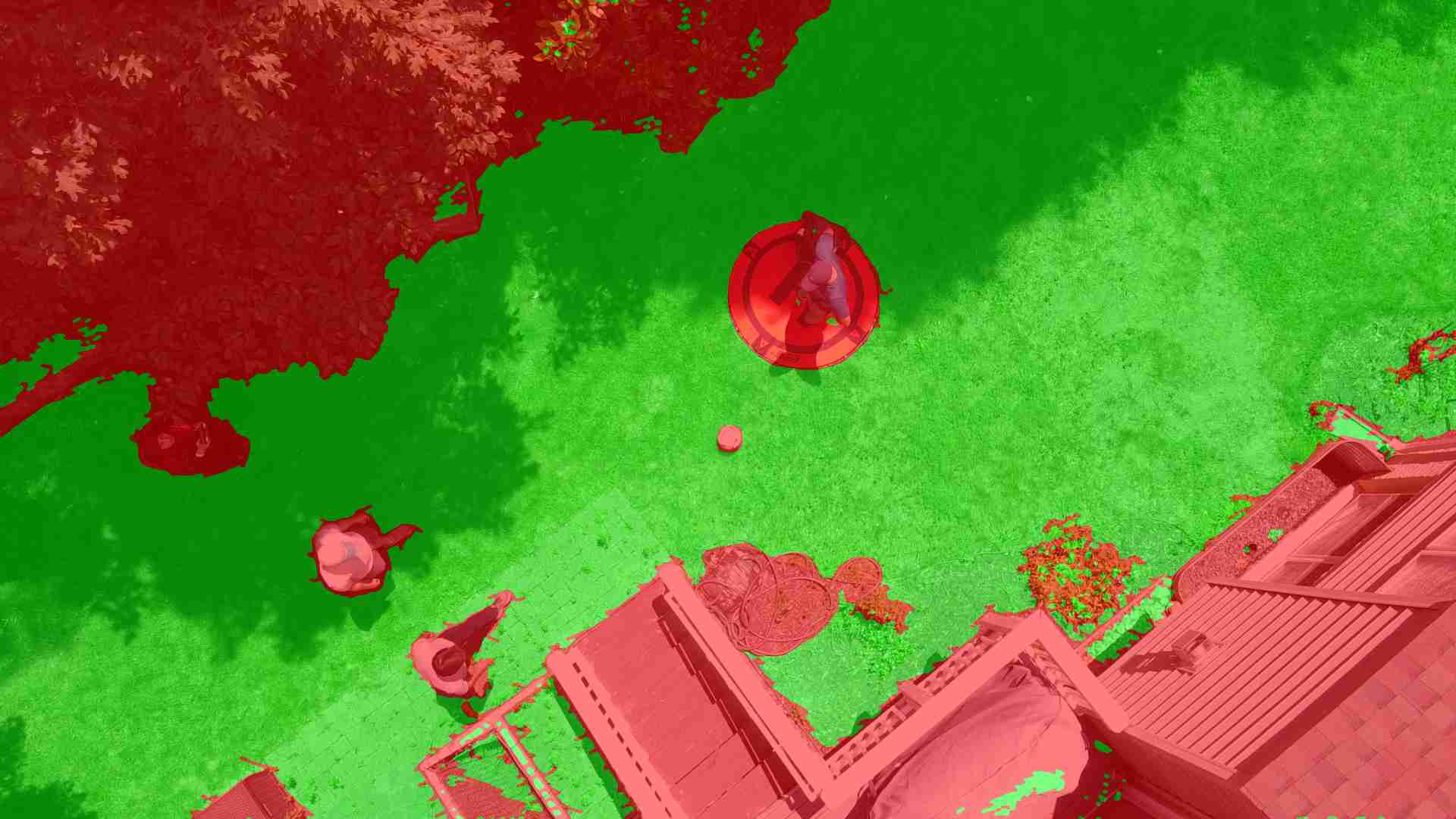} \\

\img{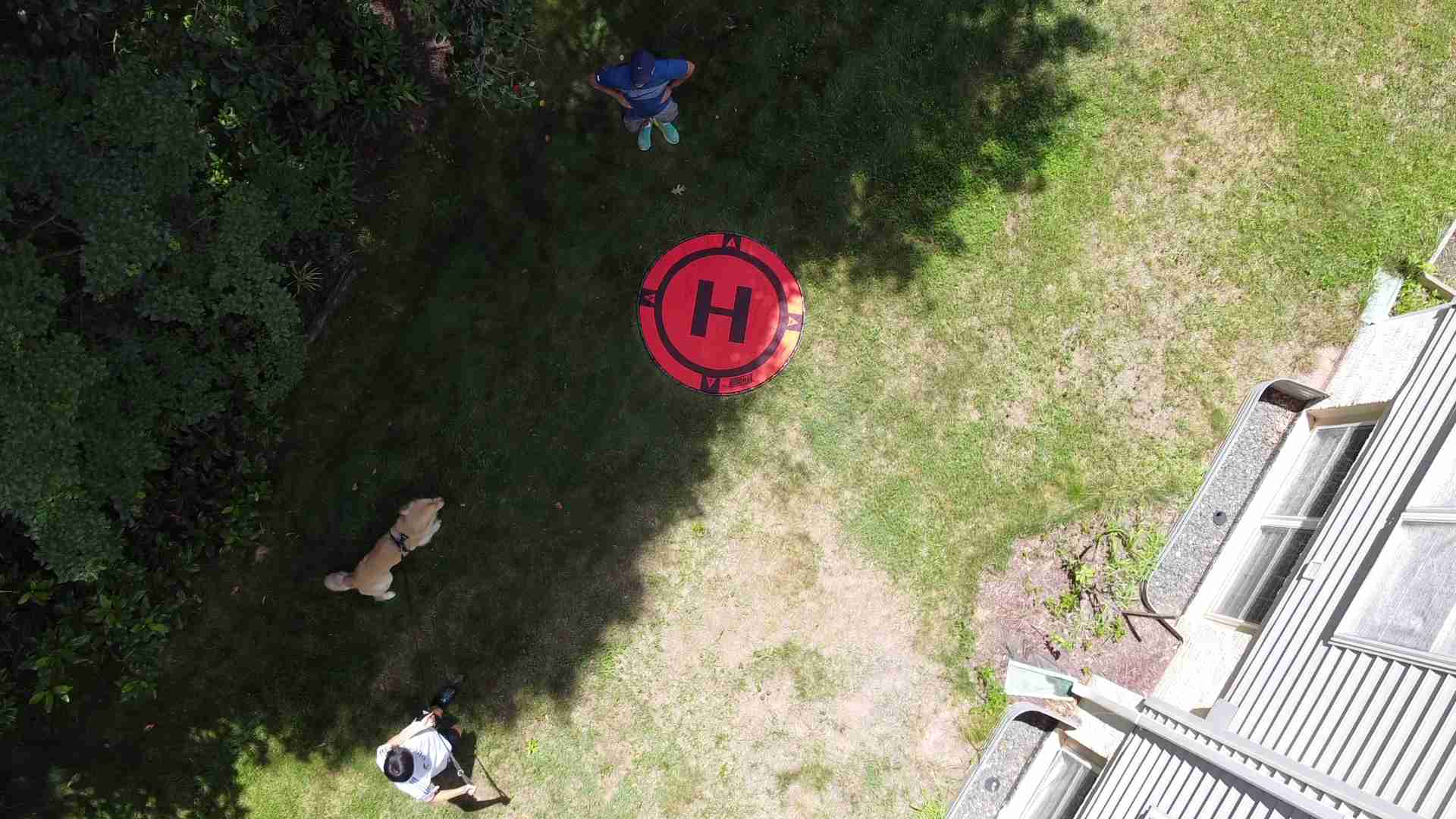} &
\img{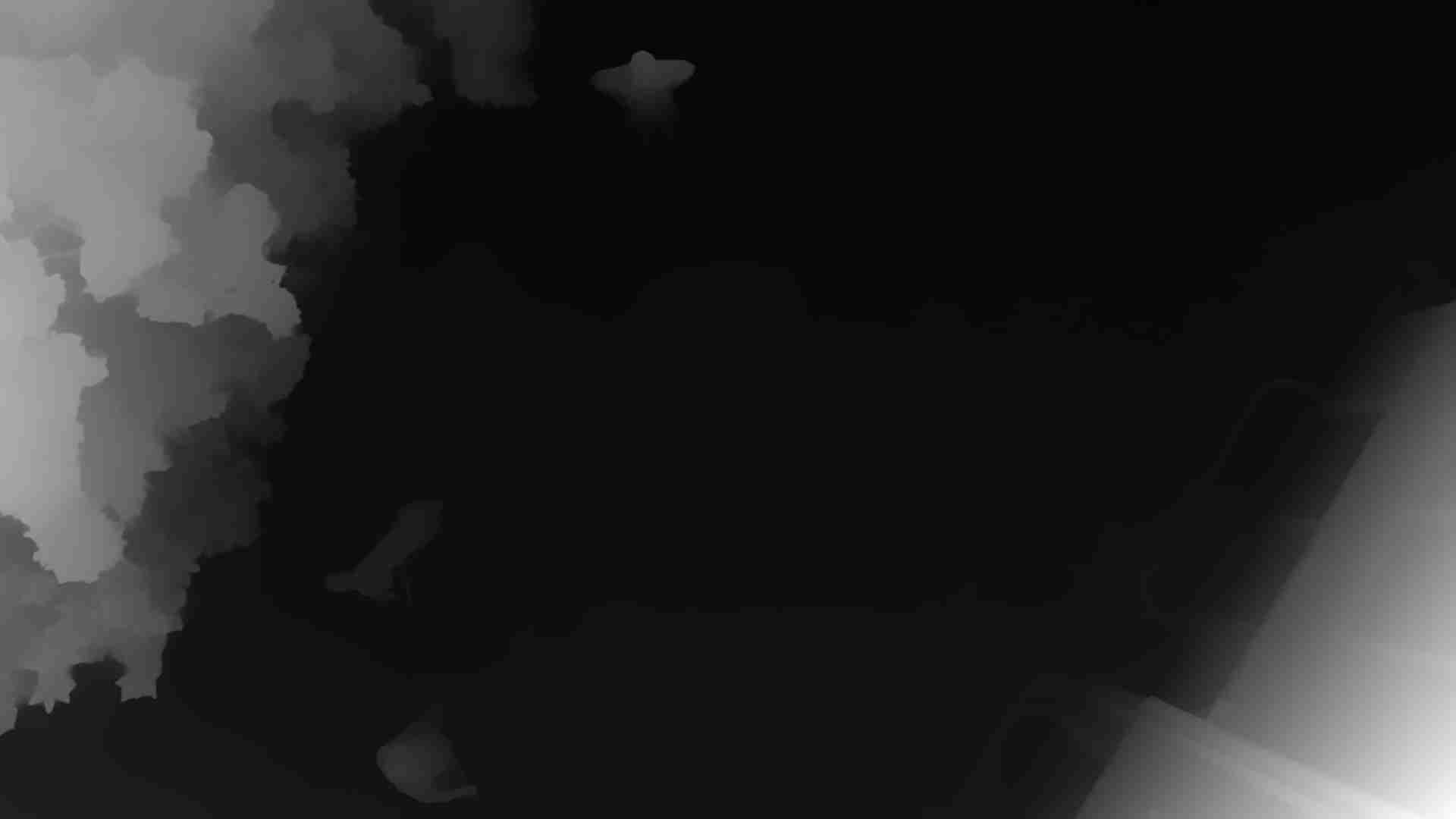} &
\img{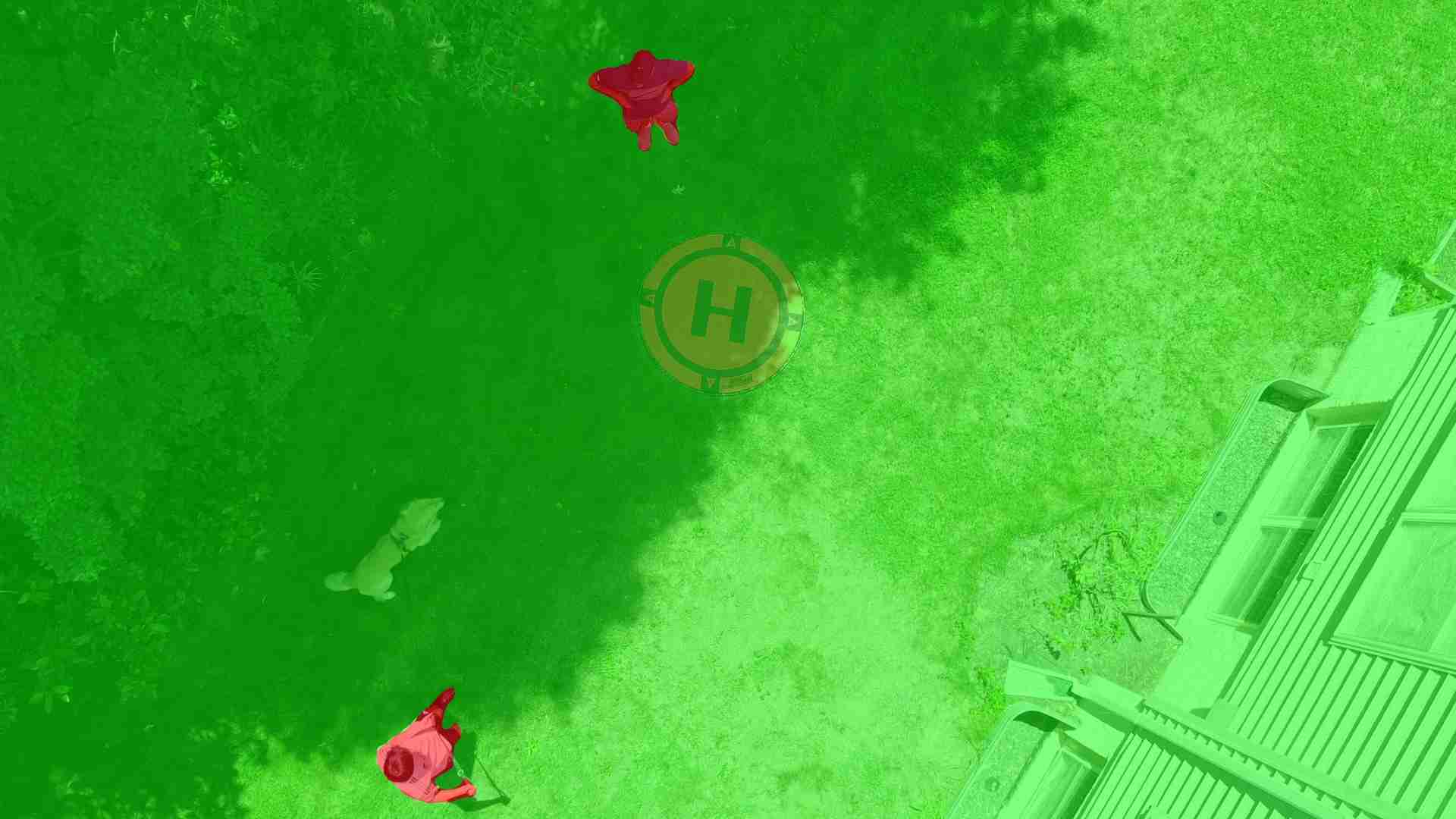} &
\img{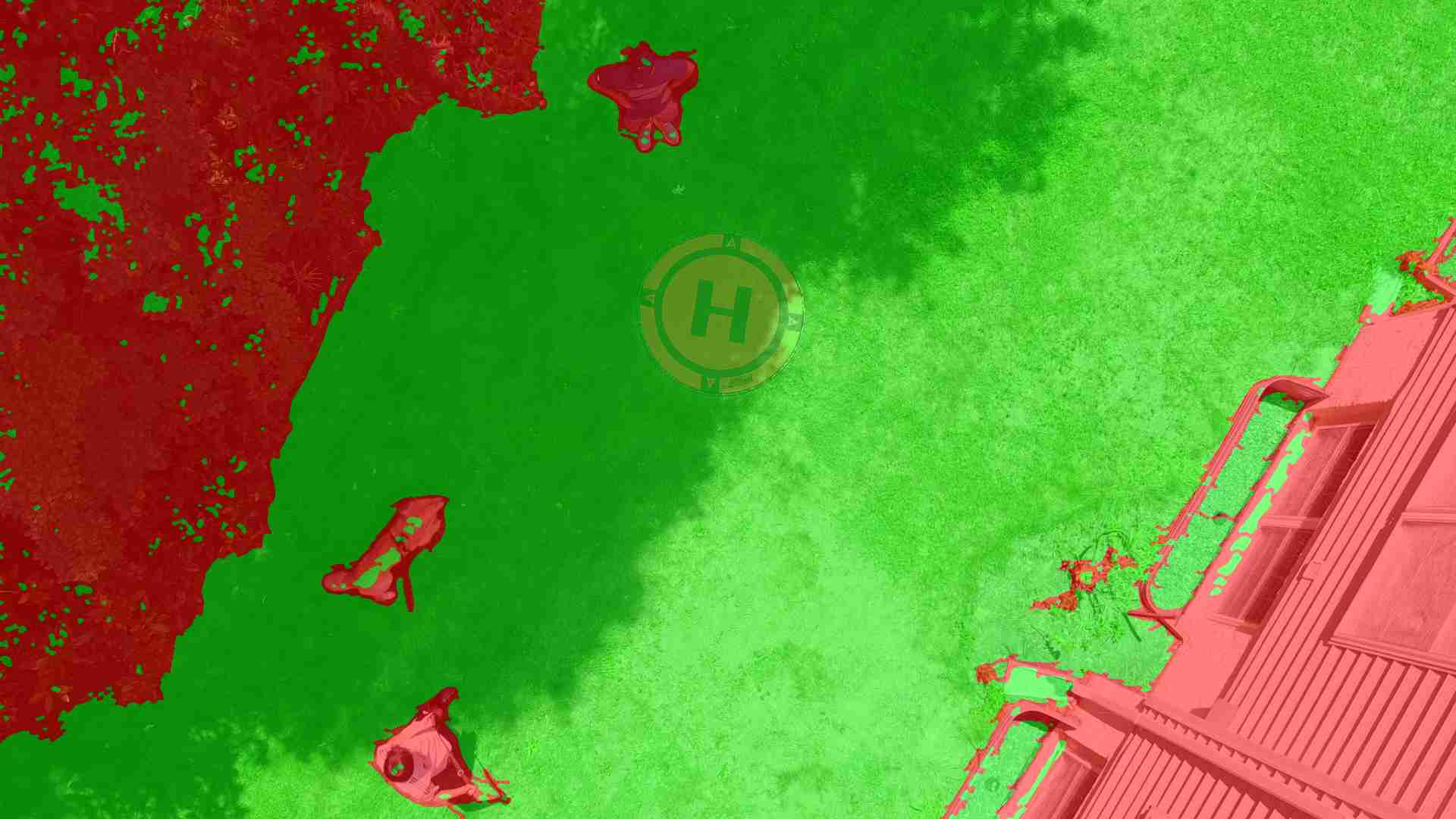} &
\img{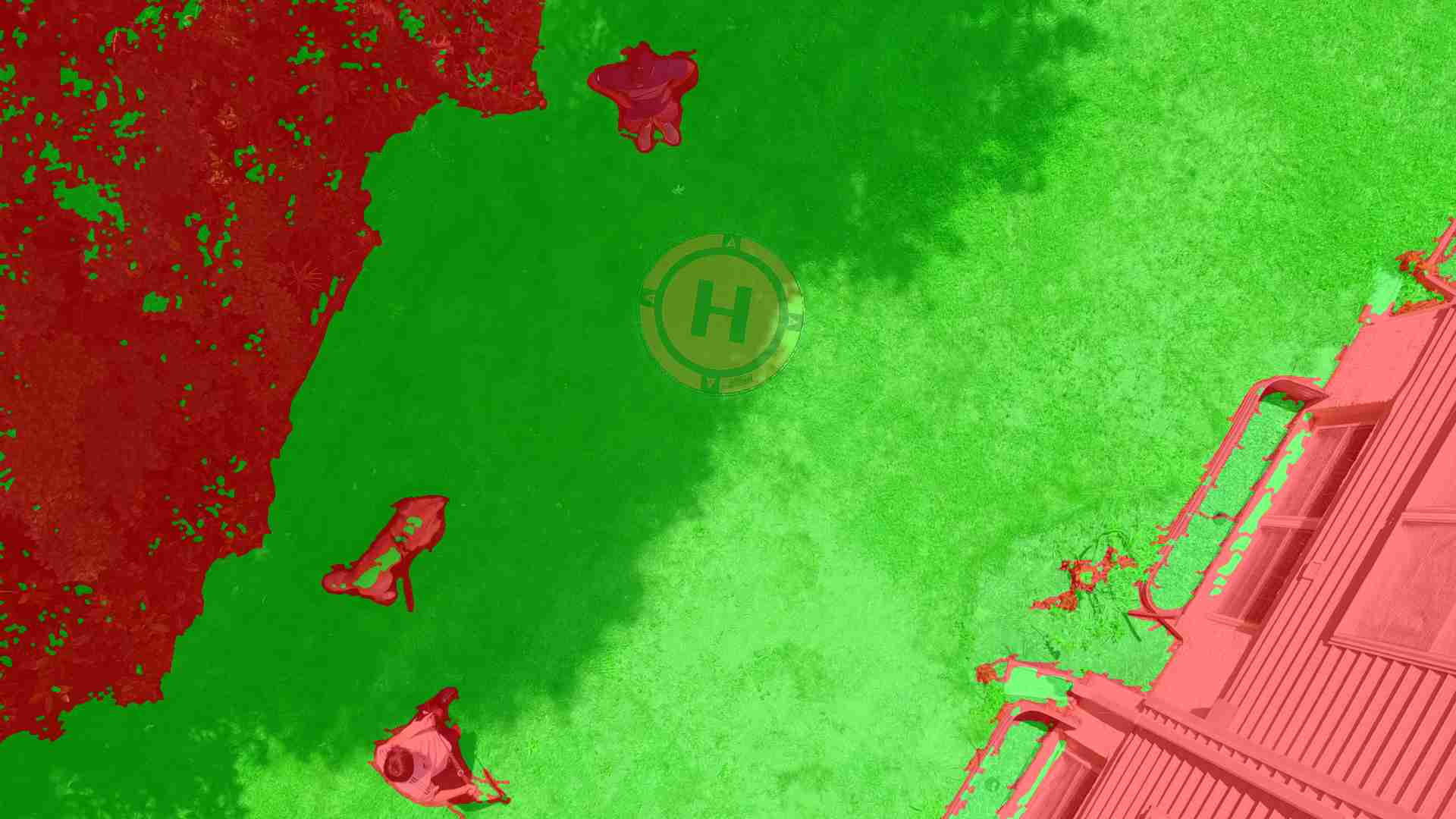} &
\img{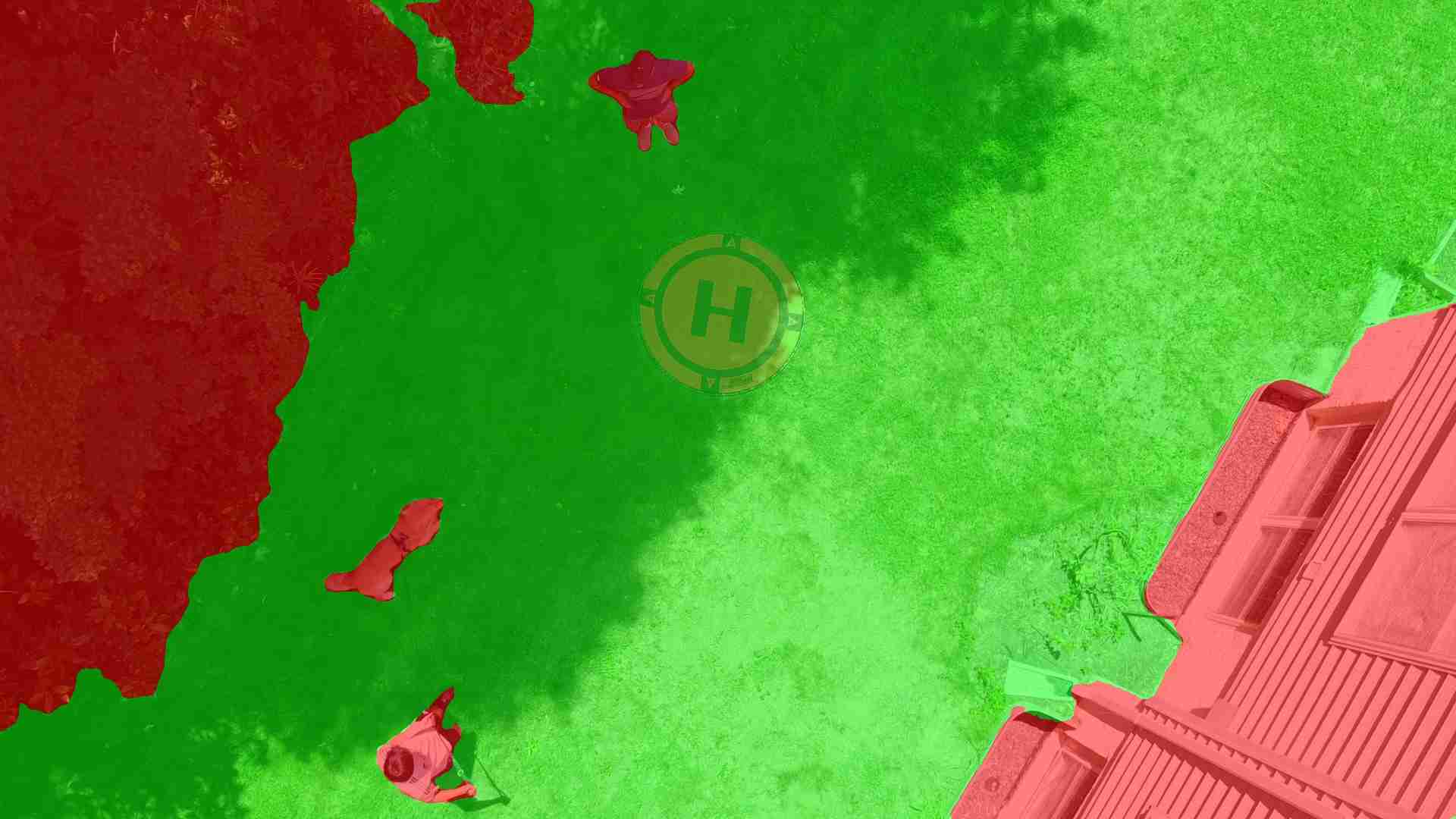} &
\img{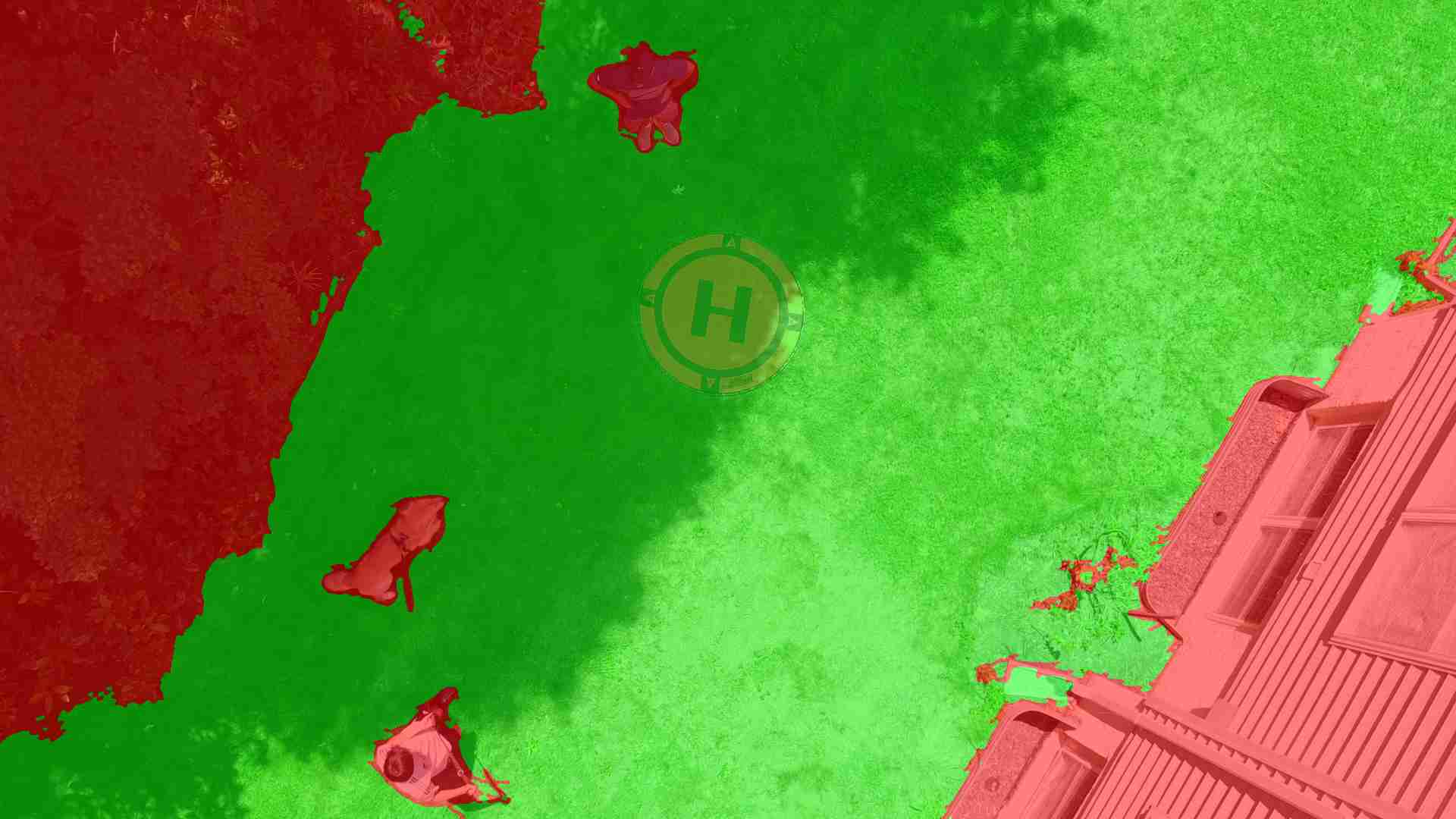} \\
\end{tabular}

\caption{Qualitative comparison of pipeline stages across three scenes.
Columns: (a) RGB input, (b) monocular depth, (c) initial DINOx mask, (d) depth-gradient mask,
(e) initial fusion (DINOx $\lor$ gradient), (f) refined DINOx (after VLM prompts), and (g) final fusion result.}
\label{fig:qual_grid_3x7}
\end{figure*}


\begin{table*}[t]
\centering
\caption{Segmentation and per-pixel detection metrics (mean~$\pm$~std). 
Best results in each column are highlighted. 
Starred entries indicate methods evaluated five times per frame.
}
\label{tab:merged-metrics}
\resizebox{\textwidth}{!}{%
\setlength{\tabcolsep}{4.5pt}
\renewcommand{\arraystretch}{1.25}
\begin{tabular}{lcccccccc}
\toprule
& \multicolumn{2}{c}{\textbf{Overlap (unsafe)}} & \multicolumn{5}{c}{\textbf{Per-pixel detection}} \\
\cmidrule(lr){2-3} \cmidrule(lr){4-8}
\textbf{Method} & \textbf{IoU} (↑) & \textbf{Dice/F1} (↑)
& \textbf{Precision} (↑) & \textbf{Recall} (↑) & \textbf{Specificity} (↑) & \textbf{Accuracy} (↑) & \textbf{Balanced Acc.} (↑) \\
\midrule
Gradient
& 0.610 $\pm$ 0.206 & 0.736 $\pm$ 0.180
& 0.965 $\pm$ 0.047 & 0.618 $\pm$ 0.205 & 0.986 $\pm$ 0.019 & 0.813 $\pm$ 0.131 & 0.802 $\pm$ 0.109 \\
RTDETR+SAM2
& 0.082 $\pm$ 0.073 & 0.144 $\pm$ 0.116
& 0.993 $\pm$ 0.021 & 0.082 $\pm$ 0.073 & 1.000 $\pm$ 0.001 & 0.592 $\pm$ 0.099 & 0.541 $\pm$ 0.037 \\
RTDETR+SAM2+Gradient
& 0.619 $\pm$ 0.200 & 0.744 $\pm$ 0.172
& 0.965 $\pm$ 0.044 & 0.628 $\pm$ 0.198 & 0.985 $\pm$ 0.019 & 0.817 $\pm$ 0.129 & 0.806 $\pm$ 0.106 \\
YOLOv8+SAM2
& 0.082 $\pm$ 0.080 & 0.143 $\pm$ 0.127
& 0.956 $\pm$ 0.204 & 0.082 $\pm$ 0.080 & \best{1.000 $\pm$ 0.000} & 0.593 $\pm$ 0.097 & 0.541 $\pm$ 0.040 \\
YOLOv8+SAM2+Gradient
& 0.619 $\pm$ 0.201 & 0.744 $\pm$ 0.173
& 0.966 $\pm$ 0.045 & 0.627 $\pm$ 0.199 & 0.986 $\pm$ 0.019 & 0.817 $\pm$ 0.129 & 0.806 $\pm$ 0.106 \\
DINOx (VisDrone category)*
& 0.081 $\pm$ 0.075 & 0.142 $\pm$ 0.119
& \best{0.997 $\pm$ 0.009} & 0.081 $\pm$ 0.075 & \best{1.000 $\pm$ 0.000} & 0.592 $\pm$ 0.100 & 0.541 $\pm$ 0.038 \\
DINOx (Flat ground category)*
& 0.560 $\pm$ 0.205 & 0.690 $\pm$ 0.170
& 0.636 $\pm$ 0.201 & 0.857 $\pm$ 0.220 & 0.477 $\pm$ 0.340 & 0.633 $\pm$ 0.196 & 0.667 $\pm$ 0.171 \\
DINOx+Gradient*
& 0.619 $\pm$ 0.200 & 0.744 $\pm$ 0.173
& 0.966 $\pm$ 0.045 & 0.627 $\pm$ 0.199 & 0.986 $\pm$ 0.019 & 0.817 $\pm$ 0.129 & 0.806 $\pm$ 0.106 \\
Vanilla VLM (o3)*
& 0.535 $\pm$ 0.115 & 0.687 $\pm$ 0.113
& 0.706 $\pm$ 0.104 & 0.684 $\pm$ 0.133 & 0.781 $\pm$ 0.068 & 0.746 $\pm$ 0.035 & 0.732 $\pm$ 0.054 \\
See\&Say (Ours)*
& \best{0.797 $\pm$ 0.120} & \best{0.880 $\pm$ 0.088}
& 0.913 $\pm$ 0.148 & \best{0.868 $\pm$ 0.044} & 0.891 $\pm$ 0.203 & \best{0.886 $\pm$ 0.106} & \best{0.880 $\pm$ 0.098} \\
\bottomrule
\end{tabular}
} 
\end{table*}

\begin{table*}[t]
\centering
\caption{Drop-zone candidate evaluation across thresholds (mean$\pm$std over frames). Best per column is highlighted. Starred entries indicate methods evaluated five times per frame.}
\label{tab:landing_candidates}
\resizebox{\textwidth}{!}{%
\setlength{\tabcolsep}{4pt}
\renewcommand{\arraystretch}{1.40}
\begin{tabular}{|l|ccc|ccc|ccc|ccc|}
\hline
\multirow{2}{*}{\textbf{Method}} 
& \multicolumn{3}{c|}{$\boldsymbol{\eta = 95\%}$} 
& \multicolumn{3}{c|}{$\boldsymbol{\eta = 90\%}$} 
& \multicolumn{3}{c|}{$\boldsymbol{\eta = 85\%}$} 
& \multicolumn{3}{c|}{$\boldsymbol{\eta = 80\%}$} \\
\cline{2-13}
& \textbf{AP} (↑) & \textbf{ROC} (↑) & \textbf{MAE} (↓)
& \textbf{AP} (↑) & \textbf{ROC} (↑) & \textbf{MAE} (↓)
& \textbf{AP} (↑) & \textbf{ROC} (↑) & \textbf{MAE} (↓)
& \textbf{AP} (↑) & \textbf{ROC} (↑) & \textbf{MAE} (↓) \\
\hline
Gradient 
& 0.741 $\pm$ 0.266 & 0.724 $\pm$ 0.224 & 0.151 $\pm$ 0.206
& 0.759 $\pm$ 0.253 & 0.795 $\pm$ 0.182 & 0.168 $\pm$ 0.178
& 0.835 $\pm$ 0.244 & 0.804 $\pm$ 0.254 & 0.174 $\pm$ 0.168
& 0.827 $\pm$ 0.239 & 0.778 $\pm$ 0.228 & 0.178 $\pm$ 0.164 \\
RTDETR+SAM2
& 0.347 $\pm$ 0.116 & 0.523 $\pm$ 0.086 & 0.410 $\pm$ 0.111
& 0.376 $\pm$ 0.128 & 0.529 $\pm$ 0.090 & 0.403 $\pm$ 0.102
& 0.444 $\pm$ 0.142 & 0.568 $\pm$ 0.107 & 0.391 $\pm$ 0.104
& 0.474 $\pm$ 0.134 & 0.538 $\pm$ 0.097 & 0.388 $\pm$ 0.106 \\
RTDETR+SAM2+Gradient
& 0.741 $\pm$ 0.266 & 0.722 $\pm$ 0.226 & 0.151 $\pm$ 0.207
& 0.759 $\pm$ 0.253 & 0.794 $\pm$ 0.183 & 0.168 $\pm$ 0.180
& 0.835 $\pm$ 0.244 & 0.804 $\pm$ 0.254 & 0.173 $\pm$ 0.168
& 0.827 $\pm$ 0.239 & 0.778 $\pm$ 0.228 & 0.177 $\pm$ 0.165 \\
YOLOv8+SAM2
& 0.347 $\pm$ 0.116 & 0.534 $\pm$ 0.074 & 0.408 $\pm$ 0.107
& 0.376 $\pm$ 0.131 & 0.535 $\pm$ 0.067 & 0.401 $\pm$ 0.099
& 0.447 $\pm$ 0.139 & 0.572 $\pm$ 0.092 & 0.390 $\pm$ 0.102
& 0.476 $\pm$ 0.127 & 0.539 $\pm$ 0.094 & 0.387 $\pm$ 0.104 \\
DINOx (VisDrone category)*
& 0.344 $\pm$ 0.117 & 0.522 $\pm$ 0.078 & 0.412 $\pm$ 0.112
& 0.372 $\pm$ 0.129 & 0.530 $\pm$ 0.076 & 0.402 $\pm$ 0.102
& 0.444 $\pm$ 0.141 & 0.567 $\pm$ 0.102 & 0.392 $\pm$ 0.103
& 0.471 $\pm$ 0.130 & 0.538 $\pm$ 0.081 & 0.387 $\pm$ 0.106 \\
DINOx (Flat ground category)*
& 0.614 $\pm$ 0.375 & 0.649 $\pm$ 0.339 & 0.122 $\pm$ 0.245
& 0.659 $\pm$ 0.367 & 0.719 $\pm$ 0.305 & 0.120 $\pm$ 0.233
& 0.731 $\pm$ 0.335 & 0.729 $\pm$ 0.345 & 0.116 $\pm$ 0.215
& 0.762 $\pm$ 0.317 & 0.729 $\pm$ 0.339 & 0.114 $\pm$ 0.206 \\
YOLOv8+SAM2+Gradient
& 0.741 $\pm$ 0.266 & 0.722 $\pm$ 0.226 & 0.151 $\pm$ 0.207
& 0.759 $\pm$ 0.253 & 0.794 $\pm$ 0.183 & 0.168 $\pm$ 0.181
& 0.835 $\pm$ 0.244 & 0.804 $\pm$ 0.254 & 0.173 $\pm$ 0.168
& 0.827 $\pm$ 0.239 & 0.778 $\pm$ 0.228 & 0.177 $\pm$ 0.165 \\
DINOx+Gradient*
& 0.741 $\pm$ 0.266 & 0.722 $\pm$ 0.226 & 0.151 $\pm$ 0.207
& 0.759 $\pm$ 0.253 & 0.795 $\pm$ 0.182 & 0.167 $\pm$ 0.178
& 0.835 $\pm$ 0.244 & 0.804 $\pm$ 0.254 & 0.173 $\pm$ 0.168
& 0.827 $\pm$ 0.239 & 0.778 $\pm$ 0.228 & 0.177 $\pm$ 0.165 \\
Vanilla VLM (o3)*
& 0.633 $\pm$ 0.192 & 0.560 $\pm$ 0.170 & 0.149 $\pm$ 0.090
& 0.675 $\pm$ 0.169 & 0.567 $\pm$ 0.167 & 0.153 $\pm$ 0.078
& 0.720 $\pm$ 0.142 & 0.563 $\pm$ 0.154 & 0.155 $\pm$ 0.066
& 0.752 $\pm$ 0.146 & 0.583 $\pm$ 0.129 & 0.160 $\pm$ 0.058 \\
See\&Say (Ours)*
& \best{0.969 $\pm$ 0.068} & \best{0.933 $\pm$ 0.131} & \best{0.036 $\pm$ 0.113}
& \best{0.978 $\pm$ 0.048} & \best{0.941 $\pm$ 0.110} & \best{0.046 $\pm$ 0.111}
& \best{0.978 $\pm$ 0.033} & \best{0.927 $\pm$ 0.102} & \best{0.051 $\pm$ 0.111}
& \best{0.982 $\pm$ 0.033} & \best{0.922 $\pm$ 0.163} & \best{0.055 $\pm$ 0.109} \\
\hline
\end{tabular}
}
\end{table*}

\begin{figure*}[t]
  \centering
  \includegraphics[width=\textwidth]{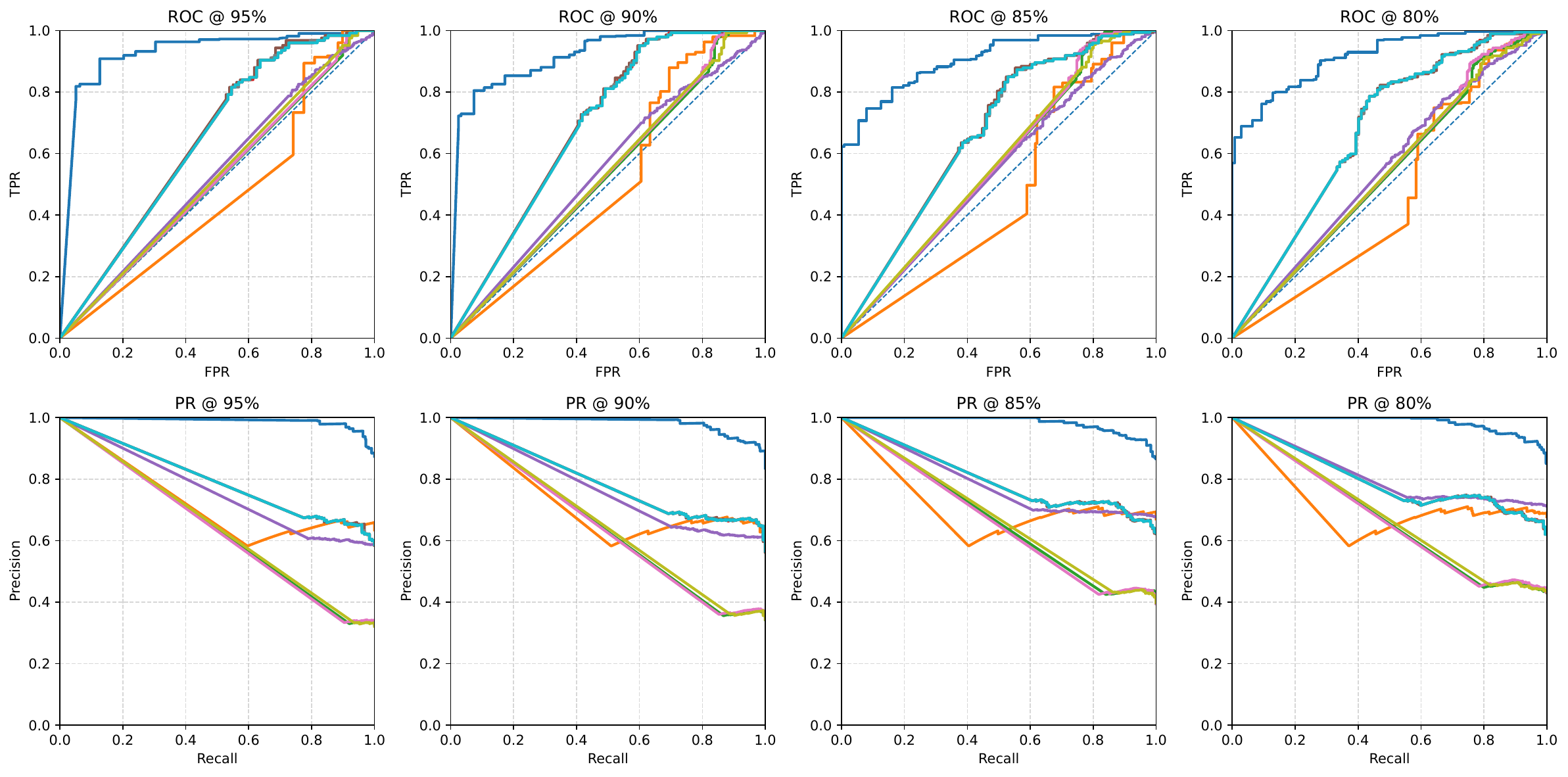}
  \vspace{0.3em}
  \includegraphics[width=\textwidth]{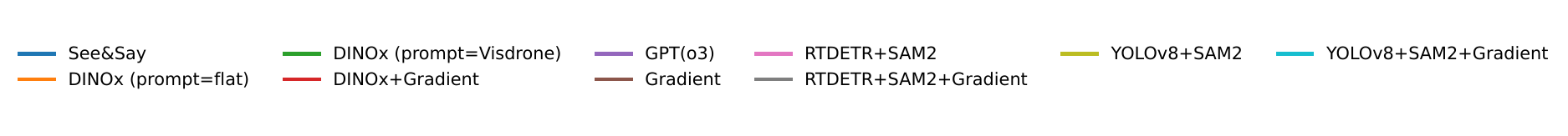}
\caption{ROC curves (top row) and Precision--Recall curves (bottom row) shown across thresholds from left to right. True positive rate (TPR) is $\frac{\text{TP}}{\text{TP}+\text{FN}}$, and false positive rate (FPR) is $\frac{\text{FP}}{\text{FP}+\text{TN}}$.}
  \label{fig:roc_pr_combined}
\end{figure*}

\subsection{Discussions}

\subsubsection{Primary Drop-Pad Safety}
Table \ref{tab:acc_reason_single} reports the success rate of three different algorithms in evaluating the safety of the primary drop-pad. We compare \textit{See\&Say} with two variants of the DINO-X algorithm.

In the first case, DINO-X is provided with the VisDrone categories as input. Here, only the ten VisDrone object classes plus the drop-pad with sign H are detected. A drop-pad is flagged as unsafe if any of the detected object masks, aside from the pad itself, overlap with the pad; otherwise, it is considered safe.

In the second case, DINO-X is given a ``complete'' category list, where all objects present in the scene are assumed to be known and passed as input prompts. The model detects masks for all such objects, and any overlap with the drop-pad is flagged as unsafe. While this setup captures more hazards than the VisDrone-only case, it still suffers from limitations: for example, scenarios where people surround the drop-pad are inherently unsafe, yet DINO-X only reasons about direct overlaps and cannot capture these higher-level risks.

By contrast, \textit{See\&Say} processes batches of five RGB frames and five depth maps jointly, which are passed to the VLM to assess whether the primary drop-pad is safe. This reasoning step is performed in parallel with safety-map generation and forms the first part of the system's output. Table \ref{tab:acc_reason_single} shows that \textit{See\&Say} outperforms both DINO-X baselines. This is expected: unlike DINO-X, the VLM leverages prior knowledge of human intentions and spatial context, allowing it to reason that close proximity of people or their motion toward the pad implies risk, even without direct overlaps. Furthermore, by considering temporal batches rather than single frame, \textit{See\&Say} exploits motion cues and temporal consistency, leading to more accurate safety assessments of the primary drop-pad. As shown in Table \ref{tab:acc_reason_single}, \textit{See\&Say} performs better when provided with a batch of five consecutive frames compared to a single frame. This result reinforces our earlier hypothesis that temporal context improves decision quality. By observing motion patterns and short-term dynamics, the VLM can better infer object trajectories, human intentions, and scene evolution.

To account for the inherent stochasticity of both VLMs and DINO-X in object detection, each model is executed five times per frame. For each run, the output is labeled as correct or incorrect, corresponding to a score of 1 or 0, respectively. The scores are first averaged across the five runs for each frame, and then aggregated over all frames to compute the final success rate.

\subsubsection{Pixel and Segmentation-wise Safety Maps}
Table \ref{tab:merged-metrics} reports the results for the safety map outputs. Each score is obtained by averaging the performance of a method across the entire set of frames, with variance reported across frames. Metrics are first computed at the frame level and then averaged over the dataset. In the case of DINO-X and gradient, the prompt list is kept consistent with the VisDrone dataset's predefined list. To ensure fairness for stochastic methods such as GPT-o3 reasoning and DINO-X (and their variants in Table \ref{tab:merged-metrics}), each algorithm is executed five times; frame-level metrics are averaged across runs before aggregating over frames. Methods evaluated in this way are denoted with a star in Table \ref{tab:merged-metrics}.

\textit{See\&Say} achieves the highest IoU (0.797), Dice/F1 (0.880), Recall (0.868), Accuracy (0.886), and Balanced Accuracy (0.880), demonstrating superior hazard coverage and stronger overlap with ground truth. Its Precision (0.913) still outperforms methods such as DINO-X (flat-ground category) and GPT-o3 reasoning, but is slightly lower than the maximum Precision reported by DINO-X (0.997). This reflects a more conservative bias, which is an acceptable trade-off in safety-critical applications where recall is paramount.

In contrast, baselines such as RT-DETR+SAM2, YOLO+SAM2, and DINO-X (VisDrone categories) report perfect specificity ($\approx 1.0$) and very high precision. However, these results are misleading: most predictions default to labeling large regions as safe, with very few false positives ($FP \approx 0$). This artificially inflates specificity and precision while failing to capture unsafe regions. Their recall is extremely low (around 0.082), meaning many unsafe areas are incorrectly classified as safe. In safety-critical tasks such as package delivery, this failure mode is hazardous, as it ignores dangerous obstacles while presenting deceptively high precision and specificity.

\subsubsection{Drop Zone Candidate Evaluation}

\textit{Table~\ref{tab:landing_candidates}} presents the results for selected landing zones across $\eta \in \{0.95, 0.90, 0.85, 0.80\}$. To evaluate landing zone selection, we first extract all candidate zones detected by each algorithm. Based on a user-defined threshold $\eta$, we then filter these candidates and retain only those whose safety ratio exceeds the threshold. The filtered candidates represent each model's final selection for the chosen threshold. 

For each candidate zone $c$, we compute the predicted and ground-truth safe ratios $\hat{s}_c$ and $s_c$ as the proportion of safe pixels within the zone, derived from the predicted and ground-truth safety maps respectively. Candidates are then assigned binary labels by thresholding at $\eta$: $\hat{y}_c = \mathds{1}[\hat{s}_c \geq \eta]$ and $y_c = \mathds{1}[s_c \geq \eta]$, from which all evaluation metrics are computed.

We evaluate the selected zones for each frame using the three landing zone metrics described earlier, and then average the results across all frames. In the case of DINO-X and gradient, the prompt list is kept consistent with the VisDrone dataset's predefined list like table \ref{tab:merged-metrics}. Again, for methods that are inherently stochastic and may vary across runs, we evaluate each frame over five runs, compute per-frame averages, and then aggregate across the dataset. Methods evaluated in this way are denoted with a star in Table~\ref{tab:landing_candidates}.

\textit{See\&Say} consistently achieves the lowest MAE and the highest AP across all thresholds, indicating both accurate safety scoring and the strongest precision--recall trade-off for safe zones. Its ROC-AUC also surpasses all baselines. Overall, \textit{See\&Say} provides the most reliable ranking and scoring of candidate zones, underscoring its suitability for autonomous drone delivery.

Figure~\ref{fig:roc_pr_combined} presents the Precision--Recall (PR) and ROC curves. These curves are constructed from pooled outputs: for each method, all candidate zones from all frames and runs are aggregated into a single prediction set prior to curve generation. This pooling provides a holistic evaluation of each method's discriminative ability. In contrast to Table~\ref{tab:landing_candidates}, which reports frame-averaged metrics, Figure~\ref{fig:roc_pr_combined} reflects dataset-level pooled evaluation. As shown, \textit{See\&Say} achieves the strongest ROC curve, demonstrating superior ability to distinguish safe from unsafe drop zones. Similarly, the PR curve for \textit{See\&Say} surpasses all baselines, highlighting its balanced performance between precision and recall across varying thresholds.

\subsubsection{Human Preference Evaluation}
\label{sec:human_eval}

\begin{figure}[t]
    \centering
    \includegraphics[width=\columnwidth]{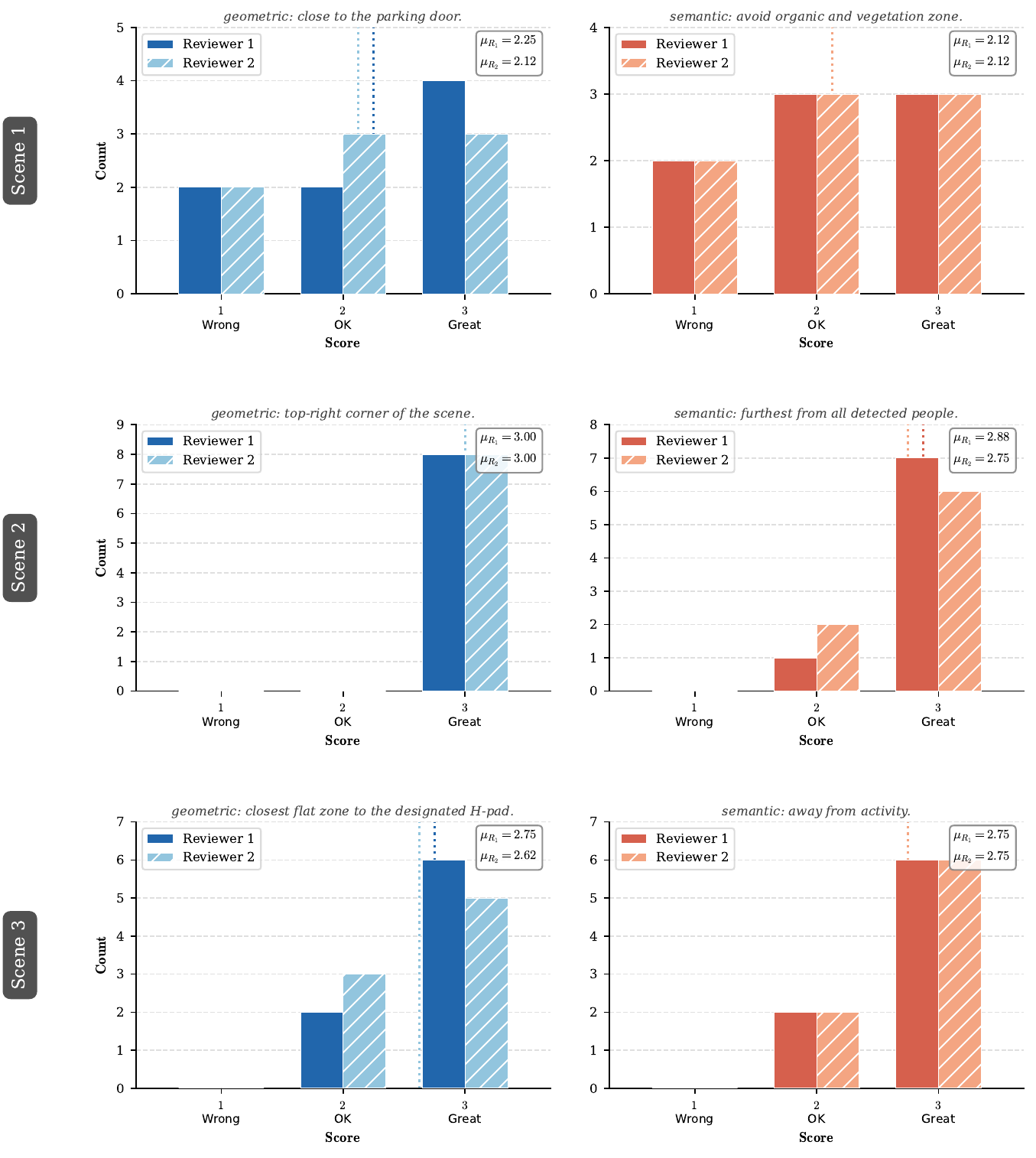}
    \caption{Score distributions for human preference evaluation across
             three scenes. Each panel shows reviewer scores (1--3) for
             geometric (left) and semantic (right) preferences.
             Dashed lines indicate $\mu_{R_1}$ and $\mu_{R_2}$.}
    \label{fig:pref_hist}
\end{figure}

To assess whether \emph{See\&Say} honors user-defined preferences when
selecting alternative drop zones, we conducted a human evaluation study
across three scenes, each evaluated under one geometric and one semantic
preference (Table~\ref{tab:human_eval} and Fig.~\ref{fig:pref_hist}).
Two independent reviewers scored each batch on a scale of 1 (preference
not satisfied) to 3 (preference well satisfied), yielding $n=8$ scored
batches per scene--preference pair.
Overall, \emph{See\&Say} achieves consistently high scores: Scene~2
receives perfect ratings ($\mu=3.00$) for both preferences from both
reviewers, with Scene~3 also performing strongly ($\mu \geq 2.62$ across
all conditions).
Scene~1 shows slightly lower scores ($\mu \approx 2.12$--$2.25$),
suggesting that tightly constrained geometric preferences (\emph{e.g.},
proximity to a specific landmark such as a parking door) are more
challenging to satisfy purely from visual reasoning.
Inter-rater agreement, measured by Cohen's $\kappa$, ranges from
substantial ($\kappa=0.810$, Scene~1 geometric) to perfect
($\kappa=1.000$, Scene~1 semantic and Scene~2 geometric), with the
lowest agreement observed for Scene~3 semantic ($\kappa=0.333$),
reflecting genuine ambiguity in interpreting \emph{``away from
activity''} in a dynamically cluttered scene.
Taken together, these results demonstrate that the VLM reasoning
component of \emph{See\&Say} effectively internalizes a diverse range
of human preferences, with stronger performance on spatially explicit
semantic preferences than on fine-grained geometric constraints.

\begin{table}[t]
  \centering
  \caption{Human preference evaluation scores (1--3) and inter-rater
           agreement (Cohen's $\kappa$) per scene and preference type.}
  \label{tab:human_eval}
  \resizebox{\linewidth}{!}{%
  \begin{tabular}{llcccr}
    \toprule
    \textbf{Scene} & \textbf{Preference} & \textbf{Avg.\ R1} & \textbf{Avg.\ R2} & $\boldsymbol{\kappa}$ & $n$ \\
    \midrule
    Scene 1 & geometric: close to the parking garage door. & 2.25 & 2.12 & 0.810 & 8 \\
     & semantic: avoid organic and vegetation zone. & 2.12 & 2.12 & 1.000 & 8 \\
    \midrule
    Scene 2 & geometric: top-right corner of the scene. & 3.00 & 3.00 & 1.000 & 8 \\
     & semantic: furthest from all detected people. & 2.88 & 2.75 & 0.600 & 8 \\
    \midrule
    Scene 3 & geometric: closest flat zone to the designated H-pad. & 2.75 & 2.62 & 0.714 & 8 \\
     & semantic: away from activity. & 2.75 & 2.75 & 0.333 & 8 \\
    \bottomrule
  \end{tabular}}
\end{table}

\section{Limitations}


While \textit{See\&Say} demonstrates promising results, it is not designed for strict real-time operation due to latency from the VLM and external API communication. However, real-time inference is not essential for the intended application: the \textit{See\&Say} reasoning module can run periodically (e.g., every 15 seconds) to capture scene updates and newly entered objects, which is sufficient for safe package-drop monitoring. For lower-latency scenarios, the system can integrate real-time open-vocabulary detectors such as YOLO-World~\cite{cheng2024yolo} for faster perception, alongside edge-accelerated inference methods ~\cite{10630963}, while reserving VLM-based reasoning for periodic safety verification. Future work will explore lightweight VLM deployment and on-device optimization to further reduce latency.

{
    \small
    \bibliographystyle{ieeenat_fullname}
    \bibliography{main}

@article{tezza2019state,
  author    = {Dante Tezza and Marvin Andujar},
  title     = {The State-of-the-Art of Human--Drone Interaction: A Survey},
  journal   = {IEEE Access},
  volume    = {7},
  pages     = {167438--167454},
  year      = {2019},
  month     = {dec},
  publisher = {IEEE},
  doi       = {10.1109/ACCESS.2019.2953900}
}

@inproceedings{tan2025vislanding,
  title     = {VisLanding: Monocular 3{D} Perception for {UAV} Safe Landing via Depth-Normal Synergy},
  author    = {Tan, Zhuoyue and He, Boyong and Ji, Yuxiang and Wu, Liaoni},
  booktitle = {Proceedings of the IEEE/RSJ International Conference on Intelligent Robots and Systems (IROS)},
  year      = {2025},
  note      = {to appear},
}

@inproceedings{vazquezlanding,
  author    = {Victoria Eugenia Vazquez-Meza and Jose Martinez-Carranza},
  title     = {Landing Zone Detection for {MAVs} Using Depth Images and Vision Transformers},
  booktitle = {Proceedings of the 15th Annual International Micro Air Vehicle Conference and Competition (IMAV 2024)},
  month     = {September},
  pages     = {162--169},
  year      = {2024},
}

@inproceedings{abdollahzadeh2022safe,
  title={Safe landing zones detection for {UAV}s using deep regression},
  author={Abdollahzadeh, Sakineh and Proulx, Pier-Luc and Allili, Mohand Said and Lapointe, Jean-Fran{\c{c}}ois},
  booktitle={2022 19th Conference on Robots and Vision (CRV)},
  pages={213--218},
  year={2022},
  organization={IEEE}
}

@article{de2025vision,
  title={Vision-Based Risk Aware Emergency Landing for {UAV}s in Complex Urban Environments},
  author={de la Torre-Vanegas, Julio and Soriano-Garcia, Miguel and Becerra, Israel and Mercado-Ravell, Diego},
  journal={arXiv:2505.20423},
  year={2025}
}

@inproceedings{tabrizian2025chain,
  title={Chain-of-Thought Flight Planner: End-to-End LLM Routing Under Wind Hazards},
  author={Tabrizian, Amin and Ghazanfari, Mahyar and Wei, Peng},
  booktitle={AIAA AVIATION FORUM AND ASCEND},
  year={2025},
  pages={3711},
}

@article{yang2024depth,
  title={Depth anything v2},
  author={Yang, Lihe and Kang, Bingyi and Huang, Zilong and Zhao, Zhen and Xu, Xiaogang and Feng, Jiashi and Zhao, Hengshuang},
  journal={Advances in Neural Information Processing Systems (NeurIPS)},
  volume={37},
  pages={21875--21911},
  year={2024}
}

@article{ren2024dinoxunifiedvisionmodel,
      title={DINO-X: A Unified Vision Model for Open-World Object Detection and Understanding}, 
      author={Tianhe Ren and Yihao Chen and Qing Jiang and Zhaoyang Zeng and Yuda Xiong and Wenlong Liu and Zhengyu Ma and Junyi Shen and Yuan Gao and Xiaoke Jiang and Xingyu Chen and Zhuheng Song and Yuhong Zhang and Hongjie Huang and Han Gao and Shilong Liu and Hao Zhang and Feng Li and Kent Yu and Lei Zhang},
      year={2024},
      journal={arXiv:2411.14347}
}

@inproceedings{kim2024weather,
  title={Weather-aware drone-view object detection via environmental context understanding},
  author={Kim, Hyunjun and Lee, Dahye and Park, Sungjune and Ro, Yong Man},
  booktitle={2024 IEEE International Conference on Image Processing (ICIP)},
  pages={549--555},
  year={2024},
}

@inproceedings{sakaino2023dynamic,
  title={Dynamic texts from {UAV} perspective natural images},
  author={Sakaino, Hidetomo},
  booktitle={Proceedings of the IEEE/CVF International Conference on Computer Vision (ICCV)},
  pages={2070--2081},
  year={2023}
}

@article{tian2025uavs,
  title={{UAV}s meet {LLM}s: Overviews and perspectives towards agentic low-altitude mobility},
  author={Tian, Yonglin and Lin, Fei and Li, Yiduo and Zhang, Tengchao and Zhang, Qiyao and Fu, Xuan and Huang, Jun and Dai, Xingyuan and Wang, Yutong and Tian, Chunwei and others},
  journal={Information Fusion},
  volume={122},
  pages={103158},
  year={2025},
  publisher={Elsevier}
}

@inproceedings{marcu2018safeuav,
  title={Safe{UAV}: Learning to estimate depth and safe landing areas for {UAV}s from synthetic data},
  author={Marcu, Alina and Costea, Dragos and Licaret, Vlad and P{\^\i}rvu, Mihai and Slusanschi, Emil and Leordeanu, Marius},
  booktitle={Proceedings of the European Conference on Computer Vision (ECCV) Workshops},
  pages={43--58},
  year={2018}
}

@inproceedings{ranftl2021vision,
  title={Vision transformers for dense prediction},
  author={Ranftl, Ren{\'e} and Bochkovskiy, Alexey and Koltun, Vladlen},
  booktitle={Proceedings of the IEEE/CVF International Conference on Computer Vision (ICCV)},
  pages={12179--12188},
  year={2021}
}

@article{chen2020robust,
  title={Robust autonomous landing of {UAV} in non-cooperative environments based on dynamic time camera-LiDAR fusion},
  author={Chen, Lyujie and Yuan, Xiaming and Xiao, Yao and Zhang, Yiding and Zhu, Jihong},
  journal={arXiv:2011.13761},
  year={2020}
}

@article{kinahan2021image,
  title={Image segmentation to identify safe landing zones for unmanned aerial vehicles},
  author={Kinahan, Joe and Smeaton, Alan F},
  journal={Irish Conference on Artificial Intelligence and Cognitive Science (AICS)},
  pages={235-247},
  year={2021}
}

@inproceedings{bultmann2021real,
  title={Real-time multi-modal semantic fusion on unmanned aerial vehicles},
  author={Bultmann, Simon and Quenzel, Jan and Behnke, Sven},
  booktitle={2021 European Conference on Mobile Robots (ECMR)},
  pages={1--8},
  year={2021},
  organization={IEEE}
}

@article{liu2022real,
  title={A real-time and multi-sensor-based landing area recognition system for {UAV}s},
  author={Liu, Fei and Shan, Jiayao and Xiong, Binyu and Fang, Zheng},
 journal   = {Drones},
  volume    = {6},
  number    = {5},
  pages     = {118},
  year      = {2022},
  month     = {May},
  publisher = {MDPI},
  doi       = {10.3390/drones6050118},
}

@inproceedings{fonder2019mid,
  title={Mid-air: A multi-modal dataset for extremely low altitude drone flights},
  author={Fonder, Michael and Van Droogenbroeck, Marc},
  booktitle={Proceedings of the IEEE/CVF Conference on Computer Vision and Pattern Recognition (CVPR) Workshops},
  pages={553--562},
  year={2019}
}

@misc{DroneDataset,
  author       = {{Institute of Computer Graphics and Vision (ICG), Graz University of Technology (TU Graz)}},
  title        = {Semantic Drone Dataset (DroneDataset)},
  howpublished = {[Online]. Available: \url{http://dronedataset.icg.tugraz.at/}},
  year         = {2019}
}

@misc{yolov8_ultralytics,
  author = {Glenn Jocher and Ayush Chaurasia and Jing Qiu},
  title = {{Ultralytics YOLOv8}},
  version = {8.0.0},
  year = {2023},
  url = {https://github.com/ultralytics/ultralytics},
  orcid = {0000-0001-5950-6979, 0000-0002-7603-6750, 0000-0003-3783-7069},
  license = {AGPL-3.0}
}

@inproceedings{redmon2016you,
  title={You only look once: Unified, real-time object detection},
  author={Redmon, Joseph and Divvala, Santosh and Girshick, Ross and Farhadi, Ali},
  booktitle={Proceedings of the IEEE Conference on Computer Vision and Pattern Recognition (CVPR)},
  pages={779--788},
  year={2016}
}

@inproceedings{zhao2024detrs,
  title={{DETR}s beat {YOLO}s on Real-Time Object Detection},
  author={Zhao, Yian and Lv, Wenyu and Xu, Shangliang and Wei, Jinman and Wang, Guanzhong and Dang, Qingqing and Liu, Yi and Chen, Jie},
  booktitle={Proceedings of the IEEE/CVF Conference on Computer Vision and Pattern Recognition (CVPR)},
  pages={16965--16974},
  year={2024}
}

@article{ravi2024sam2,
  title={SAM 2: Segment Anything in Images and Videos},
  author={Ravi, Nikhila and Gabeur, Valentin and Hu, Yuan-Ting and Hu, Ronghang and Ryali, Chaitanya and Ma, Tengyu and Khedr, Haitham and R{\"a}dle, Roman and Rolland, Chloe and Gustafson, Laura and Mintun, Eric and Pan, Junting and Alwala, Kalyan Vasudev and Carion, Nicolas and Wu, Chao-Yuan and Girshick, Ross and Doll{\'a}r, Piotr and Feichtenhofer, Christoph},
  journal={arXiv:2408.00714},

  year={2024}
}

@inproceedings{lin2014microsoft,
  title={Microsoft {COCO}: Common objects in context},
  author={Lin, Tsung-Yi and Maire, Michael and Belongie, Serge and Hays, James and Perona, Pietro and Ramanan, Deva and Doll{\'a}r, Piotr and Zitnick, C Lawrence},
  booktitle={European Conference on Computer Vision (ECCV)},
  pages={740--755},
  year={2014},
  organization={Springer}
}

@inproceedings{cao2021visdrone,
  title={VisDrone-DET2021: The vision meets drone object detection challenge results},
  author={Cao, Yaru and He, Zhijian and Wang, Lujia and Wang, Wenguan and Yuan, Yixuan and Zhang, Dingwen and Zhang, Jinglin and Zhu, Pengfei and Van Gool, Luc and Han, Junwei and others},
  booktitle={Proceedings of the IEEE/CVF International Conference on Computer Vision (ICCV)},
  pages={2847--2854},
  year={2021}
}

@inproceedings{dos2023package,
  title={Package delivery based on the leader-follower control paradigm for multirobot systems},
  author={dos Santos Cardoso, Emanuele and Bacheti, Vin{\'\i}cius Pacheco and Sarcinelli-Filho, M{\'a}rio},
  booktitle={International Conference on Unmanned Aircraft Systems (ICUAS)},
  pages={775--781},
  year={2023},
  organization={IEEE}
}

@inproceedings{slama2023risk,
  title={Risk-aware emergency landing planning for gliding aircraft model in urban environments},
  author={Sl{\'a}ma, Jakub and Herynek, J{\'a}chym and Faigl, Jan},
  booktitle={2023 IEEE/RSJ International Conference on Intelligent Robots and Systems (IROS)},
  pages={4820--4826},
  year={2023},
}

@inproceedings{moortgat2024autonomous,
  title={Autonomous {UAV} Mission Cycling: A Mobile Hub Approach for Precise Landings and Continuous Operations in Challenging Environments},
  author={Moortgat-Pick, Alexander and Schwahn, Marie and Adamczyk, Anna and Duecker, Daniel A and Haddadin, Sami},
  booktitle={2024 IEEE International Conference on Robotics and Automation (ICRA)},
  pages={8450--8456},
  year={2024},
}

@inproceedings{mitroudas2024light,
  title={Light-weight approach for safe landing in populated areas},
  author={Mitroudas, Tilemachos and Balaska, Vasiliki and Psomoulis, Athanasios and Gasteratos, Antonios},
  booktitle={2024 IEEE International Conference on Robotics and Automation (ICRA)},
  pages={10027--10032},
  year={2024},
}

@inproceedings{sinhmar2024practical,
  title={Practical and safe navigation function based motion planning of {UAV}s},
  author={Sinhmar, Himani and Greiff, Marcus and Di Cairano, Stefano},
  booktitle={2024 IEEE International Conference on Robotics and Automation (ICRA)},
  pages={12186--12192},
  year={2024},
}

@misc{zipline,
  author       = {{Zipline}},
  title        = {Zipline website},
  howpublished = {\url{https://www.zipline.com/}},
  note         = {[Online; accessed Sep. 10, 2025]},
  year         = {2025}
}

@misc{wing,
  author       = {{Wing}},
  title        = {Wing website},
  howpublished = {\url{https://wing.com/}},
  note         = {[Online; accessed Sep. 10, 2025]},
  year         = {2025}
}

@inproceedings{song2022multi,
  title={Multi-{UAV} disaster environment coverage planning with limited-endurance},
  author={Song, Hongyu and Yu, Jincheng and Qiu, Jiantao and Sun, Zhixiao and Lang, Kuijun and Luo, Qing and Shen, Yuan and Wang, Yu},
  booktitle={2022 International Conference on Robotics and Automation (ICRA)},
  pages={10760--10766},
  year={2022},
  organization={IEEE}
}

@INPROCEEDINGS{10630963,
  author={Suvizi, Ali and Subramaniam, Suresh and Lan, Tian and Venkataramani, Guru},
  booktitle={2024 IEEE Cloud Summit}, 
  title={Exploring In-Memory Accelerators and {FPGA}s for Latency-Sensitive {DNN} Inference on Edge Servers}, 
  volume={},
  number={},
  pages={1-6},
  year={2024},
  doi={10.1109/Cloud-Summit61220.2024.00007}}

@article{landis1977measurement,
  author    = {J. Richard Landis and Gary G. Koch},
  title     = {The measurement of observer agreement for categorical data},
  journal   = {Biometrics},
  volume    = {33},
  number    = {1},
  pages     = {159--174},
  year      = {1977}
}

@inproceedings{cheng2024yolo,
  title={Yolo-world: Real-time open-vocabulary object detection},
  author={Cheng, Tianheng and Song, Lin and Ge, Yixiao and Liu, Wenyu and Wang, Xinggang and Shan, Ying},
  booktitle={Proceedings of the IEEE/CVF conference on computer vision and pattern recognition (CVPR)},
  pages={16901--16911},
  year={2024}
}

@inproceedings{vu2025transformer,
  title={Transformer or {CNN}? Benchmarking Real-Time Detection Transformer and {YOLOv}8 for Small {UAS} Autonomous Landing},
  author={Vu, Can X and Ghazanfari, Mahyar and Dong, Kevin and Taye, Abenezer and Tabrizian, Amin and Wei, Peng},
  booktitle={AIAA AVIATION FORUM AND ASCEND 2025},
  pages={3521},
  year={2025}
}
}


\clearpage
\appendix
\counterwithin{table}{section}
\counterwithin{figure}{section}
\counterwithin{equation}{section}
\renewcommand{\thetable}{\Alph{section}\arabic{table}}
\renewcommand{\thefigure}{\Alph{section}\arabic{figure}}
\renewcommand{\theequation}{\Alph{section}\arabic{equation}}

\twocolumn[{
  \centering
  {\Large\bfseries See\&Say: Supplementary Material}\\[6pt]
  {\normalsize Vision--Language Guided Safe Zone Detection for
   Autonomous Package Delivery Drones}\\[12pt]
  \rule{\linewidth}{0.6pt}
  \vspace{8pt}
}]

\noindent This document provides supplementary material for the main paper,
including full implementation details, all hyperparameters, complete
VLM prompts, dataset information, and evaluation specifications.

\section{Implementation Details}
\label{sec:supp_impl}

\subsection{Software Environment}

All experiments were conducted on a Linux workstation running
Ubuntu~24.04.
The pipeline is implemented in Python~3.10 using PyTorch for
depth estimation and the OpenAI Python SDK (\texttt{openai}) for
VLM inference.
Image processing uses OpenCV~4.x and Pillow.
Depth estimation relies on the
\texttt{Depth-Anything-V2} model served locally, and open-vocabulary
detection uses the DINO-X API.

\section{Hyperparameters}
\label{sec:supp_hyperparams}

All hyperparameters were fixed across all experiments and scenes.
No scene-specific tuning was performed.

\subsection{Depth Estimation}

\begin{table}[H]
  \centering
  \caption{Depth estimation parameters.}
  \label{tab:supp_depth_params}
  \footnotesize
  \begin{tabular}{lc}
    \toprule
    \textbf{Parameter} & \textbf{Value} \\
    \midrule
    Model & Depth-Anything V2 \\
    Input resolution & Original frame size \\
    Normalization & Min--max per frame \\
    Numerical stability $\epsilon$ & $10^{-6}$ \\
    \bottomrule
  \end{tabular}
\end{table}

\subsection{Gradient-Based Flatness Filter}

The flatness mask $M^{\text{flat}}_\tau$ is computed from the
smoothed depth gradient magnitude.
Table~\ref{tab:supp_grad_params} lists all parameters.

\begin{table}[H]
  \centering
  \caption{Gradient filter hyperparameters.}
  \label{tab:supp_grad_params}
  \footnotesize
  \begin{tabular}{lcc}
    \toprule
    \textbf{Parameter} & \textbf{Symbol} & \textbf{Value} \\
    \midrule
    Gaussian smoothing $\sigma$ & $\sigma$ & 1.0 \\
    Gradient threshold & $\tau_g$ & 1.0 \\
    Local window size & $w \times w$ & $3 \times 3$ \\
    Flatness ratio threshold & $\gamma$ & 0.4 \\
    \bottomrule
  \end{tabular}
\end{table}

\noindent
A pixel at $(x,y)$ is declared \emph{flat} if at least $\gamma=0.4$
of the pixels within the $3\times3$ window centred on it satisfy
$\|\nabla D_\tau\|(u,v) < \tau_g$.
After thresholding, morphological opening and closing operations
(kernel size $5\times5$) are applied to remove small isolated
components and fill holes.

\subsection{DINO-X Open-Vocabulary Detector}

\begin{table}[H]
  \centering
  \caption{DINO-X detection parameters.}
  \label{tab:supp_dinox_params}
  \footnotesize
  \begin{tabular}{lc}
    \toprule
    \textbf{Parameter} & \textbf{Value} \\
    \midrule
    Binarisation threshold $\theta_d$ & 0.5 \\
    Initial prompt vocabulary & VisDrone-10 categories$^*$ \\
    Prompt update freq. & Once per batch (Agent~1) \\
    \bottomrule
  \end{tabular}
\end{table}

\noindent
$^*$Initial VisDrone categories:
\textit{person, pedestrian, people, bicycle, car, van, truck,
awning-tricycle, bus, motor}.

\subsection{Safety Map Fusion}

The final safety map is obtained by a pixel-wise logical union
(pessimistic fusion):
\[
  B^{\text{final}}_t = B^{\text{dinox,ref}}_t \vee U^{\text{grad}}_t.
\]
No additional weighting or learned fusion parameters are used.

\subsection{Alternative Drop Zone Selection}

\begin{table}[H]
  \centering
  \caption{Drop zone candidate selection parameters.}
  \label{tab:supp_dropzone_params}
  \footnotesize
  \begin{tabular}{@{}lcc@{}}
    \toprule
    \textbf{Parameter} & \textbf{Symbol / Key} & \textbf{Value} \\
    \midrule
    Circle radius (default) & $r$ & 100\,px \\
    Circle radius (H-pad auto) & $r$ & $\tfrac{1}{2}\sqrt{w^2+h^2}$ \\
    Minimum safe ratio & $\eta$ & 0.95 \\
    Top-$k$ candidates kept & \texttt{top\_k} & 30 \\
    Top-$n$ zones returned & \texttt{top\_n} & 3 \\
    Hex grid column spacing & $\Delta x$ & $2r$ \\
    Hex grid row spacing & $\Delta y$ & $\sqrt{3}\,r$ \\
    \bottomrule
  \end{tabular}
\end{table}

\noindent
Candidates are sorted by descending safe ratio, then descending
area, then ascending distance from the image centre before the
top-$k$ are passed to VLM Agent~2.
When no candidate satisfies $\eta=0.95$, the system saves a
\texttt{no\_candidate\_overlay.png} and returns an empty ranked list.

\subsection{VLM Agents --- Model Configuration}

Both VLM agents use the same model and API settings,
detailed in Table~\ref{tab:supp_vlm_params}.

\begin{table}[H]
  \centering
  \caption{VLM agent configuration (both agents).}
  \label{tab:supp_vlm_params}
  \footnotesize
  \begin{tabular}{lc}
    \toprule
    \textbf{Parameter} & \textbf{Value} \\
    \midrule
    Model & \texttt{o3-2025-04-16} \\
    Provider & OpenAI API \\
    Temperature & 1 (fixed, \texttt{o3}) \\
    Output format & Strict JSON \\
    Stochastic runs/frame & 5 \\
    \midrule
    \multicolumn{2}{l}{\textit{Agent 1 inputs}} \\
    \quad RGB frames & 5 (consecutive batch) \\
    \quad Depth maps & 5 (aligned to RGB) \\
    \quad Safety overlay & 1 (initial fusion) \\
    \quad Total images & 11 \\
    \midrule
    \multicolumn{2}{l}{\textit{Agent 2 inputs}} \\
    \quad RGB frames & 5 (consecutive batch) \\
    \quad Safety overlay & 1 (final, w/ candidates) \\
    \quad Candidate JSON & $\leq$30 circles \\
    \quad Total images & 6 \\
    \bottomrule
  \end{tabular}
\end{table}

\noindent
\textbf{Note on temperature.}
The \texttt{o3} reasoning model series fixes \texttt{temperature=1}
at the API level; the parameter cannot be overridden.
To account for stochasticity, each frame is evaluated five
independent times during the quantitative experiments and scores
are averaged before aggregation across the dataset
(starred entries in Tables~3--4 of the main paper).

\section{VLM Prompts}
\label{sec:supp_prompts}

We provide the complete, unmodified prompts used for both VLM
agents.
Italicised placeholders (\textit{\{...\}}) denote runtime values
substituted per batch.

\subsection{Agent 1 --- Safety Reasoning \& Prompt Refinement}

Agent~1 receives the 11-image context (5 RGB, 5 depth, 1 overlay)
and returns a structured JSON with four keys:
\texttt{landing\_pad\_safe}, \texttt{reasoning},
\texttt{future\_prediction}, and
\texttt{updated\_prompt\_list}.

\begin{tcolorbox}[
  enhanced,
  breakable,
  colback=blue!4,
  colframe=blue!35,
  fonttitle=\bfseries\small,
  title={Agent 1 System / User Prompt (Multi-Frame Mode)},
  left=6pt, right=6pt, top=4pt, bottom=4pt,
  boxrule=0.6pt
]
\small
\textbf{[No system message --- single user turn]}

\medskip
You are analyzing a package delivery drone drop safety using
5~consecutive RGB frames, their depth maps, and a
gradient-based safety overlay (green\,=\,safe, red\,=\,unsafe)
for the last frame.

\medskip
\textbf{Current DINO-X prompt list:}
\textit{\{prompt\_list\}}

\medskip
\textbf{Tasks:}
\begin{enumerate}[leftmargin=1.4em, itemsep=2pt]
  \item Determine if the landing pad is \textbf{safe} for the
    current frame (\texttt{true}/\texttt{false}).
    Decide based on the \emph{final} frame and the previous 5 frames: if there are objects
    on the landing pad, or there will be objects on the landing pad, declare unsafe, otherwise declare safe.

  \item Provide reasoning using \textbf{temporal cues} and
    \textbf{depth information}.

  \item Predict future safety (will conditions remain
    safe/unsafe?).

  \item Provide a single \textbf{updated prompt list}:
    include ALL unsafe objects/surfaces; remove safe ones
    (\emph{e.g.}\ landing pad if confirmed safe, bushes, \ldots).
    The list must reflect the most recent scene.
    Unsafe objects include any moving or static objects that are
    not flat, or are moving and not safe for a package drop.
    If the drop zone with H~sign is unsafe, also add it to
    the updated list.
    Provide the most complete prompt list for the unsafe zones.
    Avoid ambiguous prompts.
    \textbf{Rule:} streets and rooftops are always unsafe;
    bushes and grass are safe as long as they are free of objects
    and flat.
    Each entity must be specific and detectable, \emph{e.g.}\
    \textit{person, black asphalt road, white soccer ball, tree,
    white stairs, brown rooftop, \ldots}
    For \emph{people}, avoid additional details.
    For roads, stairs, decks, and ambiguous objects, specify the
    object.
    Include your reasoning for each hazardous prompt in the
    \texttt{reasoning} field.
\end{enumerate}

\medskip
\textbf{Output strictly in JSON} with keys:
\texttt{landing\_pad\_safe},
\texttt{reasoning}
(includes reasoning for choosing unsafe objects),
\texttt{future\_prediction},
\texttt{updated\_prompt\_list}
(only text prompts such as \textit{rooftop, street road, person,
landing pad with H}).

\medskip
\textbf{[Followed by 11 images: 5$\times$(RGB, depth) + overlay]}
\end{tcolorbox}

\vspace{4pt}
\noindent
\textbf{Single-frame ablation variant.}
For the single-frame ablation reported in Table~1
of the main paper, only the final RGB frame, its depth map,
and the overlay are provided (3 images total),
and the temporal reasoning instruction is replaced with a
strict rule: set \texttt{landing\_pad\_safe = false}
\emph{only} if objects are visible \emph{inside} the landing
pad area; no motion inference is permitted.

\begin{tcolorbox}[
  enhanced,
  breakable,
  colback=blue!4,
  colframe=blue!35,
  fonttitle=\bfseries\small,
  title={Agent 1 Prompt --- Single-Frame Ablation Mode},
  left=6pt, right=6pt, top=4pt, bottom=4pt,
  boxrule=0.6pt
]
\small
You are evaluating \textbf{package drop safety} for a drone.

\medskip
\textbf{Inputs:}
\begin{itemize}[leftmargin=1.4em, itemsep=1pt]
  \item ONE RGB frame (the final frame)
  \item Its depth map
  \item A safety overlay for the same frame
    (green\,=\,safe, red\,=\,unsafe)
\end{itemize}

\textbf{Current DINO-X prompt list:} \textit{\{prompt\_list\}}

\medskip
\textbf{Task (STRICT RULES):}
\begin{enumerate}[leftmargin=1.4em, itemsep=2pt]
  \item Determine whether the primary landing pad with an `H'
    marking is safe for a drop.
    Set \texttt{landing\_pad\_safe = false} \textbf{only} if
    you can see any object(s) \emph{inside} the landing pad area.
    Otherwise, set \texttt{landing\_pad\_safe = true}.
    If you cannot locate the landing pad, set it to \texttt{null}
    and explain.
  \item \texttt{reasoning}: 1--3 short sentences describing what
    you see \emph{on} the pad.
  \item \texttt{future\_prediction}: one sentence (may be empty).
  \item \texttt{updated\_prompt\_list}: if safe, return only
    clearly unsafe objects in this frame; if unsafe, also include
    \textit{landing pad with H}.
    Keep prompts concrete and detectable.
\end{enumerate}

\medskip
\textbf{Output STRICT JSON} with keys:
\texttt{landing\_pad\_safe},
\texttt{reasoning},
\texttt{future\_prediction},
\texttt{updated\_prompt\_list}.

\medskip
\textbf{[Followed by 3 images: RGB, depth, overlay]}
\end{tcolorbox}

\subsection{Agent 2 --- Preference-Guided Drop Zone Ranking}

Agent~2 receives 6 images (5 RGB frames + 1 annotated safety
overlay with candidate circles drawn) together with a compact
JSON listing all candidate circles and the user preference string.
It returns a ranked list of up to \texttt{top\_n}=3 candidates.

\begin{tcolorbox}[
  enhanced,
  breakable,
  colback=orange!5,
  colframe=orange!45,
  fonttitle=\bfseries\small,
  title={Agent 2 System Prompt},
  left=6pt, right=6pt, top=4pt, bottom=4pt,
  boxrule=0.6pt
]
\small
You are selecting circular landing zones for a drone from indexed
candidates.
\textbf{PRIORITIZE user's preference over safe ratio where they
conflict.}

Return \textbf{STRICT JSON ONLY}:
\begin{lstlisting}
{
  "ranked": [
    {
      "index": <int>,
      "reason": "<1-2 sentences>"
    },
    ...  // up to N entries
  ]
}
\end{lstlisting}
\end{tcolorbox}

\begin{tcolorbox}[
  enhanced,
  breakable,
  colback=orange!5,
  colframe=orange!45,
  fonttitle=\bfseries\small,
  title={Agent 2 User Prompt},
  left=6pt, right=6pt, top=4pt, bottom=4pt,
  boxrule=0.6pt
]
\small
\textbf{User preference:} \textit{\{user\_pref\_text\}}

\medskip
Select top \textit{\{top\_n\}} indices by preference
(ties broken by higher \texttt{safe\_ratio}).
Return \textbf{STRICT JSON ONLY}.

\medskip
\textbf{Candidates} (normalized coordinates and safe ratios):
\begin{lstlisting}
[
  {
    "index": 0,
    "cx_norm": <float>,
    "cy_norm": <float>,
    "r_norm_w": <float>,
    "r_norm_h": <float>,
    "safe_ratio": <float>
  },
  ...  // up to top_k = 30
]
\end{lstlisting}

\medskip
\textbf{[Followed by 6 images: 5 RGB frames + annotated overlay
with all candidate circles indexed]}
\end{tcolorbox}

\noindent
\textbf{Heuristic fallback.}
If the VLM response cannot be parsed or returns fewer than
\texttt{top\_n} valid indices, the remaining slots are filled
by the heuristic ranking (descending safe ratio, then area,
then proximity to image centre).

\section{Dataset Details}
\label{sec:supp_dataset}

\subsection{Recording Setup}

Videos were captured using a consumer-grade RGB camera mounted
on a drone at low altitude ($\sim$3--5\,m above ground) in
hover mode.
All three scenes were recorded in residential suburban areas
of Fairfax, VA, USA, encompassing both front-yard and
back-yard environments.
Each scene contains a physical drop pad marked with a
white H~symbol.

\subsection{Scene Descriptions}

\begin{table}[H]
  \centering
  \caption{Dataset scene summary.}
  \label{tab:supp_dataset}
  \begin{tabularx}{\linewidth}{lX}
    \toprule
    \textbf{Scene} & \textbf{Description} \\
    \midrule
    Scene~1 &
      Front yard with a paved driveway, grass area, garden hose,
      wooden deck, and 5 people moving unpredictably.
      The landing pad is on the driveway.
      Challenging due to dense human activity around and on the pad.\\
    \addlinespace
    Scene~2 &
      Back yard with open grass, a wooden deck, rooftop visible
      in the background, and maximal human presence.
      The landing pad is placed on the grass. \\
    \addlinespace
    Scene~3 &
      Back yard with mixed surface types (grass, pavement,
      flower beds), stairs, and periodic human, and dog activity.
      The landing pad is partially obstructed in several batches.
      Tests robustness to partial occlusion and cluttered geometry.\\
    \bottomrule
  \end{tabularx}
\end{table}

\subsection{Annotation Protocol}

Out of 840 frames, 120 frames (24 batches $\times$ 5 frames/batch) were
annotated by a single human expert using polygon-based
segmentation tools.
Ground-truth safety maps were produced by labelling all
\emph{unsafe} pixels, including:
non-flat surfaces (stairs, decks, rooftops, garden hoses),
dynamic objects (people, balls),
and the occupied landing pad where applicable.
The annotator was blind to model predictions during labelling.

\subsection{Batch Construction}

Frames are sampled at a fixed stride of 29 frames
($\approx 1$\,s at 30\,fps) to balance temporal coverage
with computational cost.
Each batch of 5 consecutive samples therefore spans
$\approx 5$\,s of real time, providing meaningful motion
context for the VLM.

\section{Evaluation Details}
\label{sec:supp_eval}

\subsection{Stochastic Method Evaluation}

Methods involving stochastic components (VLM-based and
DINO-X with open-vocabulary prompts) are evaluated five times
per frame.
Frame-level metrics are averaged across the five runs before
being aggregated over the dataset.
These methods are marked with a star ($\star$) in
Tables~3--4 of the main paper.

\subsection{Safety Map Metrics}

Let $\hat{B}$ and $B^*$ denote the predicted and ground-truth
binary unsafe masks respectively.
All pixel-level metrics (IoU, Dice/F1, Precision, Recall,
Specificity, Accuracy, Balanced Accuracy) are computed on the
\emph{decision frame} $I_t$ (the fifth frame of each batch)
and then averaged over all 24 batches.

For safety maps, IoU and Dice/F1 measure how well predicted unsafe masks overlap with ground-truth unsafe regions. Precision indicates how often predicted unsafe pixels are truly unsafe, while Recall captures how many unsafe pixels are correctly detected. Specificity measures how well safe regions are preserved. Accuracy provides an overall correct classification rate, and Balanced Accuracy equally weights Recall (unsafe detection) and Specificity (safe detection), which is critical for imbalanced safe/unsafe distributions.

\subsection{Drop Zone Metrics}

For drop zone selection, MAE quantifies the error between predicted and ground-truth safety ratios within candidate zones. AP evaluates the ability to identify truly safe zones with high precision across thresholds, while ROC-AUC reflects how reliably methods rank safe versus unsafe zones independent of the chosen threshold. These metrics are derived from both predicted and ground-truth safety maps. For each candidate zone $c$, we compute the predicted and ground-truth safe ratios:
\[
    \hat{s}_c = \text{pred.\ safe ratio}, \quad s_c = \text{gt.\ safe ratio},
\]
where the safe ratio is defined as the proportion of safe pixels relative to all pixels within the zone.
Each candidate is then assigned a binary ground-truth label: if $s_c \geq \eta$, the zone is labeled as $1$ (safe); otherwise, it is labeled as $0$ (unsafe). Predicted labels are similarly obtained by thresholding the predicted safe ratio:
\[
    \hat{y}_c = \mathbf{1}[\hat{s}_c \ge \eta], \quad y_c = \mathbf{1}[s_c \ge \eta],
\]
where $\hat{y}_c$ denotes the predicted label and $y_c$ the ground-truth label for zone $c$.

Using these labels and scores, we evaluate the performance of each method on the filtered candidate zones. Table~\ref{tab:supp_metrics} summarizes all metrics and their mathematical definitions.

\begin{table}[t]
\centering
\small
\renewcommand{\arraystretch}{1.6}
\caption{Evaluation metrics for safety maps and drop zone selection.}
\label{tab:supp_metrics}
\resizebox{\columnwidth}{!}{%
\begin{tabular}{lll}
\toprule
\textbf{Category} & \textbf{Metric} & \textbf{Formula} \\
\midrule
\multirow{7}{*}{Safety Map}
& IoU & $\tfrac{TP}{TP+FP+FN}$ \\
& Dice/F1 & $\tfrac{2TP}{2TP+FP+FN}$ \\
& Precision & $\tfrac{TP}{TP+FP}$ \\
& Recall & $\tfrac{TP}{TP+FN}$ \\
& Specificity & $\tfrac{TN}{TN+FP}$ \\
& Accuracy & $\tfrac{TP+TN}{TP+TN+FP+FN}$ \\
& Balanced Acc. & $\tfrac{\text{Recall}+\text{Spec.}}{2}$ \\
\midrule
\multirow{3}{*}{Drop Zone}
& MAE & $\tfrac{1}{N}\sum_{c=1}^N|\hat{s}_c-s_c|$ \\
& AP & $\int_0^1 \text{Prec}(r)\, d(\text{Rec}(r))$ \\
& ROC-AUC & $\int_0^1 TPR(t)\, d(FPR(t))$ \\
\bottomrule
\end{tabular}
}
\end{table}

\section{Human Preference Evaluation Protocol}
\label{sec:supp_human_eval_protocol}

The preference evaluation described in Section~4.3.4 of the
main paper was conducted using a structured two-reviewer protocol.

\subsection{Review Package Construction}

For each of the three scenes, two preferences were selected:
one geometric and one semantic (Table~\ref{tab:supp_pref_list}).
The pipeline was run independently for each scene--preference
pair using a \texttt{prefs\_json} file specifying the same
preference string for all 8 batches within that run.

\begin{table}[H]
  \centering
  \caption{Preferences used in the human evaluation study.}
  \label{tab:supp_pref_list}
  \begin{tabularx}{\linewidth}{llX}
    \toprule
    \textbf{Scene} & \textbf{Type} & \textbf{Preference string} \\
    \midrule
    Scene~1 & Geometric & \textit{Close to the parking door} \\
    Scene~1 & Semantic  & \textit{Avoid organic and vegetation zones} \\
    \midrule
    Scene~2 & Geometric & \textit{Top-right corner of the scene} \\
    Scene~2 & Semantic  & \textit{Furthest from all detected people} \\
    \midrule
    Scene~3 & Geometric &
      \textit{Closest flat zone to the designated H-pad} \\
    Scene~3 & Semantic  & \textit{Away from activity} \\
    \bottomrule
  \end{tabularx}
\end{table}

\subsection{Scoring Procedure}

Each reviewer independently inspected, for every batch:
(i) the five RGB input frames,
(ii) the final safety overlay with ranked candidate circles
drawn on it (\texttt{ranked\_on\_overlay.png}), and
(iii) the VLM-generated reasoning for each ranked zone.
Reviewers assigned an integer score from the following rubric:

\begin{center}
\footnotesize
\begin{tabular}{cl}
  \toprule
  \textbf{Score} & \textbf{Meaning} \\
  \midrule
  1 & Completely wrong / irrelevant to preference \\
  2 & Partially correct / acceptable \\
  3 & Very well suited to preference \\
  \bottomrule
\end{tabular}
\end{center}

\noindent
The two reviewers scored independently with no communication
during the annotation phase.
Inter-rater agreement is reported as unweighted Cohen's
$\kappa$ (Table~4 of the main paper).

\subsection{Cohen's $\kappa$ Interpretation}

Following standard conventions~\cite{landis1977measurement},
$\kappa$ values are interpreted as:
$<0.20$~poor, $0.21$--$0.40$~fair, $0.41$--$0.60$~moderate,
$0.61$--$0.80$~substantial, $>0.80$~almost perfect.
The lowest agreement in our study ($\kappa=0.333$,
Scene~3 semantic) is categorised as \emph{fair},
attributable to genuine ambiguity in interpreting
``away from activity'' across dynamically varying batches
rather than a systematic disagreement on system quality.

\section{Additional Qualitative Results}
\label{sec:supp_qualitative}

Figure~1 of the main paper illustrates the pipeline output for
one representative batch.
Here we note additional qualitative observations across the
three scenes:

\begin{itemize}[leftmargin=1.4em, itemsep=3pt]

  \item \textbf{VLM prompt refinement:}
    Across all scenes, Agent~1 consistently expanded the
    initial VisDrone-10 vocabulary to include scene-specific
    hazards not present in the original list, such as
    \textit{coiled garden hose}, \textit{wooden deck and
    railing}, \textit{soccer ball}, and
    \textit{tree canopy}.
    This demonstrates the zero-shot generalisation capability
    that fixed-vocabulary baselines cannot replicate.

  \item \textbf{Temporal reasoning:}
    In batches where a person walks onto the landing pad mid-
    sequence, Agent~1 correctly flags the pad as unsafe despite
    the person being absent in earlier frames, leveraging
    motion cues across the 5-frame window.
    Single-frame evaluation (Table~1 of the main paper) confirms
    that removing temporal context reduces success rate from
    0.975 to 0.958.

  \item \textbf{No-candidate batches:}
    In Batch~7 of Scene~1 (frames 004060--004176), the safety
    map labels the entire visible area as unsafe due to dense
    human activity, resulting in zero feasible circles at
    $\eta=0.95$.
    The system correctly reports no alternative zones rather
    than returning an unsafe candidate.

  \item \textbf{Preference adherence:}
    For Scene~2 with preferences ``top-right corner'' and
    ``furthest from all detected people'', both reviewers
    assigned perfect scores ($\mu=3.00$, $\kappa=1.000$),
    indicating that spatially explicit preferences are reliably
    satisfied when the safety map correctly identifies the
    relevant open area.

\end{itemize}

\section{Reproducibility Checklist}
\label{sec:supp_repro}

\begin{itemize}[leftmargin=1.4em, itemsep=3pt]
  \item[$\checkmark$] All hyperparameters reported in
    Section~\ref{sec:supp_hyperparams}.
  \item[$\checkmark$] Complete VLM prompts provided in
    Section~\ref{sec:supp_prompts}.
  \item[$\checkmark$] Dataset recording conditions described in
    Section~\ref{sec:supp_dataset}.
  \item[$\checkmark$] Evaluation protocol (stochastic averaging,
    metrics) described in Section~\ref{sec:supp_eval}.
  \item[$\checkmark$] Human evaluation scoring rubric provided in
    Section~\ref{sec:supp_human_eval_protocol}.
\end{itemize}

\end{document}